\documentclass[sigplan,nonacm]{acmart}
\usepackage{graphicx} 
\usepackage{lipsum}

\usepackage{pdfpages}
\usepackage{bm}
\usepackage{algorithm}
\usepackage{algorithmic}
\usepackage{listings}
\usepackage{subfig}
\usepackage{array}
\usepackage{enumitem}
\usepackage{xcolor}
\usepackage{caption}
\usepackage{tikz}

\usepackage{hyperref}
\makeatletter
\def\UrlAlphabet{%
      \do\a\do\b\do\c\do\d\do\e\do\f\do\g\do\h\do\i\do\j%
      \do\k\do\l\do\m\do\n\do\o\do\p\do\q\do\r\do\s\do\t%
      \do\u\do\v\do\w\do\x\do\y\do\z\do\A\do\B\do\C\do\D%
      \do\E\do\F\do\G\do\H\do\I\do\J\do\K\do\L\do\M\do\N%
      \do\O\do\P\do\Q\do\R\do\S\do\T\do\U\do\V\do\W\do\X%
      \do\Y\do\Z}
\def\UrlDigits{\do\1\do\2\do\3\do\4\do\5\do\6\do\7\do\8\do\9\do\0}
\g@addto@macro{\UrlBreaks}{\UrlOrds}
\g@addto@macro{\UrlBreaks}{\UrlAlphabet}
\g@addto@macro{\UrlBreaks}{\UrlDigits}
\makeatother
\usepackage[hyphenbreaks]{breakurl}

\definecolor[named]{ACMPurple}{cmyk}{0.55,1,0,0.15}

\newcommand{\colorcite}[1]{\textcolor{ACMPurple}{\cite{#1}}}
\newcommand{\colorref}[1]{\textcolor{ACMPurple}{\ref{#1}}}

\usepackage[]{hyperref}
\title{FastGL: A GPU-Efficient Framework for Accelerating Sampling-Based GNN Training 
at Large Scale}

\author{Zeyu Zhu$^{1,2}$ \ Peisong Wang$^{1}$ \
Qinghao Hu$^{1}$ \ Gang Li$^{3}$  \ Xiaoyao Liang$^{3}$ \ Jian Cheng$^{1,2,4,5}\ddagger$
}
\affiliation{%
  \institution{$^1$Institute of Automation, CAS}
  \institution{$^2$School of Future Technology, University of Chinese Academy of Sciences}
  \institution{$^3$Shanghai Jiao Tong University}
  \institution{$^4$AiRiA \ $^5$Maicro.ai}
  \country{}
}
\email{{zhuzeyu2021, huqinghao2014}@ia.ac.cn, gliaca@sjtu.edu.cn, liang-xy@cs.sjtu.edu.cn
}
\email{{peisong.wang, jcheng}@nlpr.ia.ac.cn}

\begin{document}

\graphicspath{{figures/}}

\begin{abstract}
  { \spaceskip=0.95\fontdimen2\font plus 0.7\fontdimen3\font minus 1.15\fontdimen4\font
    Graph Neural Networks (GNNs) have shown great superiority on non-Euclidean graph data,
  achieving ground-breaking performance on various graph-related tasks. 
  As a practical solution to train GNN on large graphs 
  with billions of nodes and edges, the sampling-based training is widely adopted
  by existing training frameworks. However, through an in-depth analysis, 
  we observe that the efficiency of existing sampling-based training frameworks 
  is still limited due to the key bottlenecks lying in all three phases of 
  sampling-based training,
  i.e.,  
  subgraph sample, memory IO, and computation.
  To this end,  
  we propose \textbf{FastGL}, a GPU-efficient
  \underline{\textbf{F}}ramework for \underline{\textbf{a}}ccelerating
  \underline{\textbf{s}}ampling-based \underline{\textbf{t}}raining of 
  \underline{\textbf{G}}NN at 
  \underline{\textbf{L}}arge scale by simultaneously optimizing all above three
  phases, 
  taking into account both GPU characteristics and graph structure.
  Specifically, by
  exploiting the inherent overlap within graph structures, FastGL develops
  the
  \textbf{Match-Reorder} strategy to 
  reduce the data traffic, which accelerates the memory IO without 
  incurring any GPU memory overhead. 
  Additionally, FastGL leverages a \textbf{Memory-Aware} computation method, 
  harnessing the GPU memory's 
  hierarchical nature to mitigate irregular data access during computation.
  FastGL further incorporates the
  \textbf{Fused-Map} approach aimed at diminishing the synchronization overhead 
  during sampling.
  Extensive experiments demonstrate that FastGL can achieve an average speedup of
  {$11.8\times$, $2.2\times$ and $1.5\times$} over
  the state-of-the-art frameworks 
  PyG, DGL, and GNNLab, respectively. Our code is 
  available at \href{https://github.com/a1bc2def6g/fastgl-ae}
  {\textcolor{ACMPurple}{https://github.com/a1bc2def6g/fastgl-ae}}.
  }
\end{abstract}
\maketitle 
\pagestyle{plain} 

\begingroup
\spaceskip=0.95\fontdimen2\font plus 0.9\fontdimen3\font minus 1.2\fontdimen4\font
\section{Introduction}

\renewcommand{\thefootnote}{$\ddagger$}
\footnotetext[4]{Corresponding author.}
\renewcommand{\thefootnote}{$*$}
\footnotetext[3]{This is the author's version of the work. 
It is posted here for personal use. Not for redistribution. 
The definitive Version of Record was published in 
29th ACM International Conference on Architectural Support for 
Programming Languages and Operating Systems, Volume 4 (ASPLOS '24), April 
27-May 1, 2024, La Jolla, CA, USA, http://dx.doi.org/10.1145/3622781.3674167}
\captionsetup{font=small}
Recently, Graph Neural Networks (GNNs) have attracted
much attention from both industry and academia due to their
superior learning and representing ability for non-Euclidean graph data.
A number of GNNs  \colorcite{kipf2016semi,xu2018powerful,
hamilton2017inductive,velivckovic2018graph} have been widely used in various
graph-related tasks, such as social network analysis \colorcite{lerer2019pytorch,fan2019graph}, 
autonomous driving \colorcite{weng2020joint,klimke2022cooperative}, 
and recommendation systems \colorcite{jin2020multi,yang2020hagerec}.

\begin{figure}[t]
  \centering
  \hspace{-0.3cm}
  \includegraphics[scale=0.43,trim=0cm 0cm 0cm 0.0cm,clip]{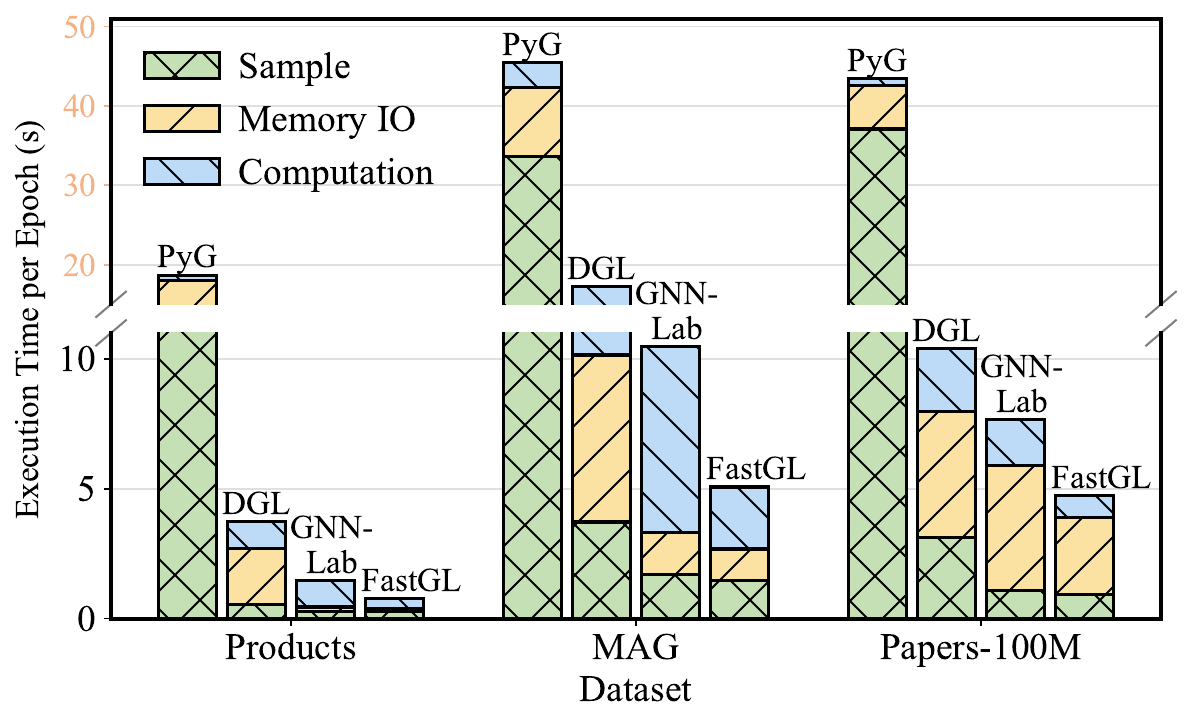}
  \caption{The execution time breakdown analysis of training GCN on various graphs using 
  different sampling-based training frameworks.
  More experiment setup is detailed in Section \colorref{experiment_setup}.}
\label{time_breakdown_intro}
\end{figure}

In practice, many real-world graphs are large-scale and associated with
rich node attributes (i.e., features). For example, the Pinterest graph \colorcite{ying2018graph} contains
over 2 billion nodes and 17 billion edges with 18TB data size.   
Such a large size makes it impossible to load the entire
graph into the GPU with only tens of gigabytes memory, which turns down proposals designed for
full-graph training on GPUs \colorcite{jia2020improving,wan2022bns}. 
A practical solution to this problem is the sampling-based GNN
training \colorcite{hamilton2017inductive,huang2018adaptive,ying2018graph}, 
which adopts a sampling algorithm
to extract subgraphs from the entire graph and trains GNNs on these sampled subgraphs.

For the sampling-based
GNN training frameworks \colorcite{fey2019fast,wang2019deep,
zheng2020distdgl,liu2023bgl,yang2022gnnlab,lin2020pagraph},
the CPU host memory is used to store graphs 
including graph topology information and nodes features,
and GPUs are responsible for model training. 
Each iteration of the sampling-based training consists of three stages: (1) \textit{sample} subgraphs and
store them in host memory, (2) load structures of subgraphs and features of sampled nodes to 
GPUs (referred to as \textit{memory IO} phase),
(3) launch the forward and backward \textit{computations} for GNN training. 
While the sampling-based training is ingenious, it comes with a significant drawback - 
the high overhead of data traffic and sampling. This is due to the large number of 
mini-batches that need to be trained \colorcite{fey2019fast,wang2019deep}, 
which seriously slows down the training process and hinders
GNNs' real-world applications on large-scale graphs. 
Although a spectrum of  
frameworks \colorcite{fey2019fast,wang2019deep,
zheng2020distdgl,liu2023bgl,yang2022gnnlab,lin2020pagraph,zheng2022bytegnn} have been proposed
to expedite this process, through our in-depth analyses in 
Figure \colorref{time_breakdown_intro}, we observe that they still 
cannot {unlock} the acceleration potential due to the following three 
performance bottlenecks:

\textbf{Insufficient support for fast sampling.} 
First, existing frameworks often perform graph sampling on the CPU, 
which is extremely time-consuming due to the poor parallelism.
For example, 
PyG \colorcite{fey2019fast} 
spends up to 97\% 
of the overall training 
time to sample on CPU.
Second,
in the sample phase, 
the global ID of each sampled node in the raw graph needs to be converted to a local ID 
in the sampled graph (this process is referred 
to as \textit{ID map}) to avoid loading all nodes features to the GPU memory. 
Although DGL \colorcite{wang2019deep} utilizes
the thread parallelism of GPU to accelerate the sample phase, the {vast volume} 
of thread synchronizations to 
obtain the local IDs still
incurs considerable latency overhead into the sample phase, which
occupies up to 38\% time.

\textbf{Failing to accelerate the memory IO on large-scale graphs.} 
{Since the bandwidth between host 
memory and GPU memory is low (e.g., 32GB/s of 
PCIE 4.0 with 16 channels), 
the memory IO phase
dominates the overall training time (up to 77\%)
and significantly impedes performance boosting. This phenomenon is exacerbated
as the graph size increases.}
PaGraph \colorcite{lin2020pagraph} and GNNLab \colorcite{yang2022gnnlab} both regard 
a portion of GPU memory as a software-controlled
cache to reduce the data traffic between the CPU and GPU. However,
the device memory in GPU is a scarce resource and there is little or no extra memory
to be used as a cache especially in the large-scale graph scenario, which makes these methods
inefficient.

\textbf{Irregular memory access during the computation.} For real-world graphs, 
the connections are extremely sparse and 
unstructured \colorcite{li2021gcnax,hu2020open,hamilton2017inductive}, 
and hence inducing great irregularity into 
the computation phase when accessing data 
(e.g., features and edge weights) 
according to edges. 
The irregularity results in poor hit rates (4.41\%/19.6\% on average) of the L1/L2 cache 
on GPU, which seriously 
limits the bandwidth available to the computing units,
thereby hampering the GPU performance 
and decelerating the computation. Previous works overlook 
optimizations on memory accesses, making the computation phase another bottleneck.

To tackle the above challenges, 
we focus on accelerating all phases of the sampling-based GNN 
training in this paper by 
taking into account both GPU characteristics and graph structure 
{under the single machine with multiple GPUs scenario}.
For the biggest bottleneck (except in PyG), i.e., memory IO, 
we first reveal that there are great overlap across different sampled subgraphs, 
stemming from the complex graph structures.
This motivates us to design a greedy strategy of \textbf{Match-Reorder} to maximally
reuse the overlapping nodes between mini-batches, which reduces the magnitude of data traffic 
between CPU and GPU without imposing any GPU memory cost.
Furthermore, to mitigate the irregularity in the computation, we introduce 
the \textbf{Memory-Aware} computation approach that 
capitalizes on the disparities in bandwidth across different GPU memory levels. 
By adjusting the memory access pattern,  
we enhance the overall bandwidth utilization and GPU performance, 
consequently speeding up the computation phase.
Finally, 
we develop the \textbf{Fused-Map} sampling method to fuse the necessary operations in the ID map process,
thereby significantly reducing the volume of thread synchronizations and 
expediting the sample phase.

We integrate the above design ideas into \textbf{FastGL}, a GPU-efficient \underline{\textbf{F}}ramework tailored
for \underline{\textbf{a}}ccelerating
\underline{\textbf{s}}ampling-based \underline{\textbf{t}}raining of 
\underline{\textbf{G}}NN on 
\underline{\textbf{L}}arge-scale  
graphs, which substantially accelerates 
the sampling-based GNN training.

In summary, the contributions of this paper are as follows:
\begin{itemize}[leftmargin=*,topsep=0pt]
  \item We conduct a breakdown analysis of the sampling-based GNN training on 
  large-scale 
  graphs and identify three main performance bottlenecks.\vspace{-0.1cm}
  To mitigate the biggest bottleneck, 
  i.e., memory IO, we propose the \textbf{Match-Reorder} strategy to maximally reduce 
  the data traffic without inducing any GPU memory overhead.
  \item 
  To accelerate the computation phase, we put forward the \textbf{Memory-Aware} 
  computation approach, 
  which redesigns the memory access pattern to increase the GPU performance by
  exploiting different levels of memory in GPU.
  \item We propose the \textbf{Fused-Map} sampling method to avoid the thread synchronizations and 
  expedite the sample phase by 
  performing the ID map process in a fused mechanism.
  \item We implement the proposed framework \textbf{FastGL} 
  equipped with the above three techniques and conduct extensive experiments 
  with various GNN models and large-scale graph datasets.
  On average, FastGL can improve the training efficiency by 
  {$11.8\times$, $2.2\times$ 
  and $1.5\times$} over the state-of-the-art
  PyG \colorcite{fey2019fast}, DGL \colorcite{wang2019deep}, and GNNLab \colorcite{yang2022gnnlab}, 
  respectively.
\end{itemize}

\section{Background and Related Work}
\subsection{Graph Neural Networks.}
Graph Neural Networks
(GNNs) \colorcite{kipf2016semi,xu2018powerful,hamilton2017inductive,velivckovic2018graph} 
are deep learning models aiming at addressing graph-related tasks in an end-to-end manner.
Given a graph $G=(V,E)$, which consists of $|E|$ 
edges and $|V|$ nodes, and each node is associated
with a feature vector. 
In each GNN layer, the forward pass contains 
two phases: the aggregation and the update. 
In the aggregation phase, the target node $u$ collects weighted information from its
neighbors (referred to as source nodes) following the graph structure and generates the hidden features as follows:
\begin{equation}
  \label{aggregate_information}
\setlength{\abovedisplayskip}{3pt}
\bm{h}_{u}^{k+1} = \sum\limits_{v \in N(u)}w_{uv}^{k}\cdot \bm{x}_{v}^{k} \ \text{,}
\setlength{\belowdisplayskip}{3pt}
\end{equation}  
where $k$ denotes the \textit{k-th} layer of GNN, $N(u)$ represents the neighbors of the 
node $u$, $\bm{x}_{v} \in \mathbb{R} ^{d}$ is the feature vector
of node $v$, 
$w_{uv} \in \mathbb{R} $ is the weight 
and $\bm{h}_{u}$ is the aggregated hidden features of node $u$. 
Then, the hidden features of each node are transformed to the new features 
as the output of the current layer in the update phase as follows: 
\begin{equation}
  \label{update_information}
\setlength{\abovedisplayskip}{3pt}
\bm{x}_u^{k+1}=\text{Update}^{k+1}(\bm{h}_u) \ \text{.}
\setlength{\belowdisplayskip}{3pt}
\end{equation}  
A GNN model learns a low-dimensional embedding for each node 
by recursively performing the above process,
which would be used in various downstream tasks
\colorcite{kipf2016semi,hamilton2017inductive,fan2019graph}. 

\begin{figure}[t]
  \vspace{-0.2cm}
  \centering
  \includegraphics[scale=0.55,trim=6.5cm 6.2cm 8.7cm 5cm,clip]{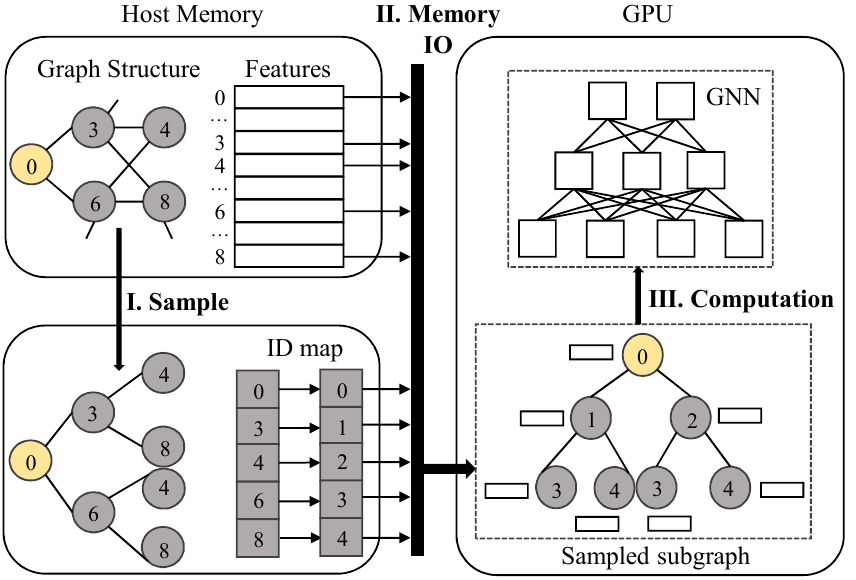}
  \vspace{-0.1cm}
  \caption{An example of the sampling-based training
  in a 2-layer GNN on the training node 0. The sampling algorithm
  uniformly selects two neighbors for each node.}
  \vspace{-0.3cm}
\label{sample_process}
\end{figure}

\subsection{Sampling-based GNN Training.}
There are two categories of
training paradigms adopted in existing GNN systems: full-batch training and sampling-based 
training. Full-batch training
loads the entire graph into GPUs for training \colorcite{kipf2016semi}, such as NeuGraph \colorcite{ma2019neugraph}, 
ROC \colorcite{jia2020improving}, 
and FlexGraph \colorcite{wang2021flexgraph}. 
Unfortunately, for large-scale
graphs, 
such an approach is limited by the GPU memory capacity. Therefore, we focus 
on the sampling-based training in this paper.

The sampling-based approach
(also referred to as mini-batch training) \colorcite{chen2018fastgcn,
hamilton2017inductive,huang2018adaptive,zeng2019graphsaint} splits the
training nodes into multiple mini-batches and samples the subgraphs according 
to these mini-batches and then conducts
GNN training on subgraphs iteratively. 
Training on one mini-batch
is called an iteration, which 
consists of three
phases as shown in Figure \colorref{sample_process}, i.e., \textbf{sample}, \textbf{memory IO} 
and \textbf{computation}.
In general, the sample phase is divided into two steps. (1) \textbf{sample subgraph:} 
Starting from each node in a
mini-batch, the input subgraph is sampled according to a user-defined sample algorithm. 
(2) \textbf{ID map:}  
To update the subgraph structure,
the global ID of each sampled node in the raw graph needs to be
converted to a local ID (consecutive and start from 0) in the sampled graph.
After sampling,
the features of sampled
nodes and subgraph structures are loaded from host memory to GPU 
memory. 
Finally, GPU launches the 
computation for GNN training 
on the sampled subgraph.

\subsection{GNN Training Frameworks.}

In recent years, two categories of GNN training frameworks are also
developed upon existing deep learning 
frameworks for efficient GNN training,
i.e., full-batch training 
framework \colorcite{ma2019neugraph,jia2020improving,wang2021flexgraph,
thorpe2021dorylus,wan2022bns,wan2023scalable} 
and sampling-based training framework \colorcite{zheng2020distdgl,wang2019deep,
fey2019fast,zhu12aligraph,gandhi2021p3,huan2022t,yang2023betty}.
Constrained by GPU memory, full-batch training frameworks cannot 
scale efficiently to large-scale graphs, thus we focus on the 
presentation of sampling-based training frameworks in this section. 

PyG \colorcite{fey2019fast} integrates with PyTorch \colorcite{paszke2019pytorch} 
to provide a message-passing API for GNN training. However, sampling by CPUs, 
PyG cannot scale well to 
large graphs. 
DGL \colorcite{wang2019deep} accelerates the sample process by utilizing GPUs, which has 
much higher parallelism than CPUs. 
However, the enormous thread synchronizations
on GPUs incur heavy latency overhead 
to the sample.
And the vast volume of data traffic
between host memory and GPU memory further limits the training efficiency. 
Much worse, current frameworks \colorcite{wang2019deep,fey2019fast,zheng2020distdgl} 
ignore the problem of poor
cache hit rates caused by the extreme irregularity of the graph structure, 
which leads to the computation phase being another performance bottleneck. 

To overcome the above problems, some more efficient sampling-based GNN training frameworks 
are proposed \colorcite{yang2022gnnlab,lin2020pagraph,
wang2021gnnadvisor,liu2023bgl,zheng2022bytegnn,cai2023dsp}. PaGraph \colorcite{lin2020pagraph} 
and GNNLab \colorcite{yang2022gnnlab}
both regard a portion of GPU memory as a cache and
propose the corresponding cache policy to reduce
the data traffic between the CPU and GPU. However,
GPU memory is a scarce resource, and there
is often no extra space on the GPU to be used as a cache, especially when the
sampled subgraphs are large. Dedicated for full-batch training, 
GNNAdvisor \colorcite{wang2021gnnadvisor} preprocesses the graph
according to its properties and 
proposes a 2D workload management to accelerate the computation. 
However, for the sampling-based training, the sampled subgraphs must be 
preprocessed in each iteration, 
incurring severe preprocessing overhead. There are also some 
works \colorcite{jangda2021accelerating,pandey2020c,kaler2022accelerating} 
that aim to accelerate the sample phase 
based on GPU or CPU. {Unfortunately, sampling on CPUs
\colorcite{kaler2022accelerating} do not 
utilize GPU parallelism, and \colorcite{jangda2021accelerating,pandey2020c} 
overlook the latency overhead induced by the numerous thread synchronizations.}

\section{Motivation}

\subsection{Rethinking the Acceleration of Memory IO}

\label{rethink_memory}
To identify the bottleneck of the sampling-based GNN training, we conduct an 
execution time breakdown analysis of its three phases,
i.e., sample, memory IO, and computation
on Products dataset of GCN \colorcite{kipf2016semi} and GIN \colorcite{xu2018powerful} model, as 
depicted in Figure \colorref{time_breakdown}. 
`Naive' denotes that the time is obtained by running the model with DGL \colorcite{wang2019deep}.
We observe that by
consuming up to 77\% of the training time,
the memory IO dominates the overall training process.
This is primarily due to the vast volume of 
data transferred from host memory to GPU memory, a challenge further 
compounded in large-scale graphs by the limited bandwidth 
(e.g., 32GB/s of PCIE 4.0 with 16 channels).

\begin{figure}[t]
  \vspace{-0.2cm}
  \centering
  \hspace{-0.3cm}
  \includegraphics[scale=0.43,trim=0cm 0cm 0cm 0.0cm,clip]{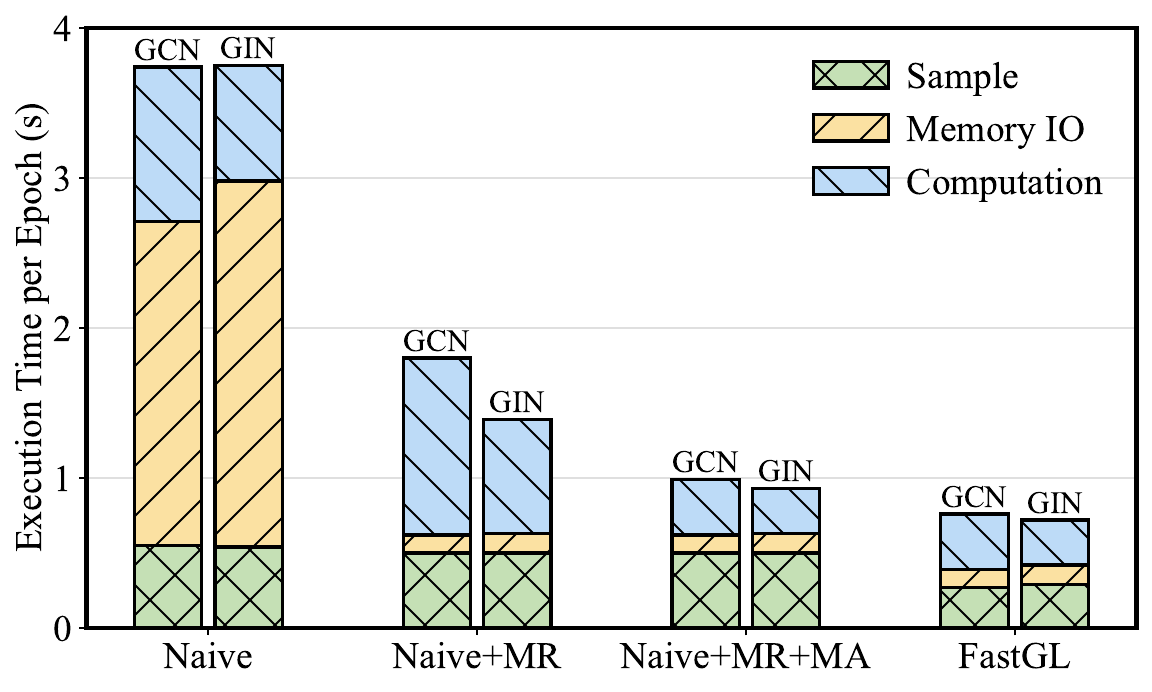}
  \vspace{-0.1cm}
  \caption{The execution time breakdown analysis on Products dataset of GCN and GIN on 
  two NVIDIA RTX 3090 GPUs. The average time is generated from 20 epochs.}
\label{time_breakdown}
\vspace{-0.3cm}
\end{figure}

\begin{table}[t]
  \caption{The remaining GPU memory when running a 3-layer GCN on different graphs based on DGL. Measured on 
  one 3090 GPU with 24GB memory, the batch size is 8000, and the hidden size is 256.}
  \label{left_memory}
  \begin{center}
      \begin{tabular}{ccccc}
        \toprule[1.5pt]
                     Graph & Reddit  & Products  & MAG      &\begin{tabular}[c]{@{}c@{}}Papers-100M \end{tabular} \\ \midrule[1pt]
                     \begin{tabular}[c]{@{}c@{}}Left   \\ Memory\end{tabular}& 13GB & 11GB   & 520MB  &1GB  \\ \bottomrule[1.5pt]
      \end{tabular}
  \end{center}
  \vspace{-0.2cm}
\end{table}

{Many existing works focus on resolving the bottleneck of 
memory IO \colorcite{lin2020pagraph,yang2022gnnlab,liu2023bgl,song2022rethinking}, whose key idea
is to  
regard a portion of GPU memory as a
software-controlled cache, and perform various optimizations on the cache policy to cache 
nodes features more efficiently. }
Upon cache hit, the data traffic between CPU and GPU can be saved. 
However, as presented in Table \colorref{left_memory},
the GPU memory is a scarce resource and 
there is little or no left memory
to be used as a cache because the sampled subgraphs are large, which is caused by 
the neighbor explosion problem \colorcite{chiang2019cluster,zeng2019graphsaint}.
In such a scenario, the previous works can only cache a small (even no) fraction  
of nodes features on GPU due to 
the memory constraints and they are no longer effective in reducing data traffic. 
For example, the hit rate of PaGraph is less than 20\% on MAG.
Although naively reducing the mini-batch size leaves 
more GPU memory available, this proportionally increases the number of mini-batches to be trained, 
which significantly 
offsets the benefits gained from memory I/O optimization.

Thanks to the complex topology in the large-scale graph, i.e., one node 
often connects with multiple nodes and may be sampled in different subgraphs, 
there is a high degree of node overlap (up to 96\%) 
between different subgraphs. 
This motivates us to propose the Match-Reorder (MR) strategy 
where Match reduces the data traffic by reusing the overlapping nodes
without imposing any GPU memory cost and Reorder maximizes the reuse.

\begin{table}[t]
  \caption{The hit rates of L1/L2 cache and the achievable performance of 3090 GPU 
  in the forward process of 
  the aggregation phase. 
  }
  \label{cache_hit}
  \begin{center}
    \begin{tabular}{ccccc}
      \toprule[1.5pt]
      Graph     & Reddit & Products & MAG    & Papers-100M \\ \midrule[1pt]
      L1   Cache & 3.34\% & 5.11\%   & 4.92\% & 4.25\%      \\
      L2   Cache & 24.6\% & 18.3\%   & 15.7\% & 19.6\%      \\ \hline
      GFLOP/s & 340 & 397   & 380 & 401      \\ \bottomrule[1.5pt]
      \multicolumn{5}{l}{\small * Measured by the NVIDIA Nsight Compute 2023.1.}
      \end{tabular}
  \end{center}
  \vspace{-0.2cm}
\end{table}

\begin{table}[t]
  \caption{The statistics of different memory levels in 3090 GPU.}
  \label{bandwidth_level}
  \begin{center}
    \begin{tabular}{ccccp{1cm}<{\centering}}
      \toprule[1.5pt]
                & \begin{tabular}[c]{@{}c@{}}L1   \\ Cache\end{tabular} & \begin{tabular}[c]{@{}c@{}}Shared   \\ Memory\end{tabular} & \begin{tabular}[c]{@{}c@{}}L2   \\ Cache\end{tabular} & \begin{tabular}[c]{@{}c@{}}Global   \\ Memory\end{tabular} \\ \midrule[1pt]
      Bandwidth & $\sim$12TB/s                                          & $\sim$12TB/s                                               & 3$\sim$5TB/s                                          & 938GB/s                                                    \\ \hline
      Capacity & \multicolumn{2}{c}{128KB (per SM)}                                                                                     & 6MB                                          & 24GB                                                    \\ \bottomrule[1.5pt]  
    \end{tabular}
  \end{center}
\end{table}

\subsection{Opportunity in the Computation Acceleration}
\label{memory_aware_motivation}
As shown in Figure \colorref{time_breakdown} (`Naive+MR'), 
after optimizing the memory IO, 
the computation phase renders the primary performance
bottleneck due to the 
the great irregularity of memory accesses
induced by the extremely sparse
connections between nodes during the aggregation.
Table \colorref{cache_hit} presents the L1/L2 cache hit rate and the achievable GPU performance 
during 
the aggregation phase. 
The low hit rate limits the
bandwidth attainable to the computing units and results in 
the achievable GPU performance being much lower than the theoretical peak performance 
(29155GFLOP/s of 3090 GPU),  
incurring severe latency overhead into the computation.

Fortunately, in addition to the L1/L2 cache, 
there are multiple memory levels on GPU 
with widely varying bandwidth, as shown in Table \colorref{bandwidth_level}.
Intuitively, the more data is stored in the memory with a higher bandwidth, 
the higher overall bandwidth is available to the computing units.
However, limited by the memory capacity, only a fraction of data can be stored in the higher bandwidth 
memory.
Accordingly, we first analyze 
the memory access pattern in the aggregation and then propose
the Memory-Aware (MA) method, which
prioritizes the storage of frequently
accessed data in higher-bandwidth memory
to increase the overall bandwidth and 
accelerate the computation phase.

\begin{figure}[t]
  \centering
  \includegraphics[scale=0.45,trim=6.4cm 3.2cm 6.3cm 4.6cm,clip]{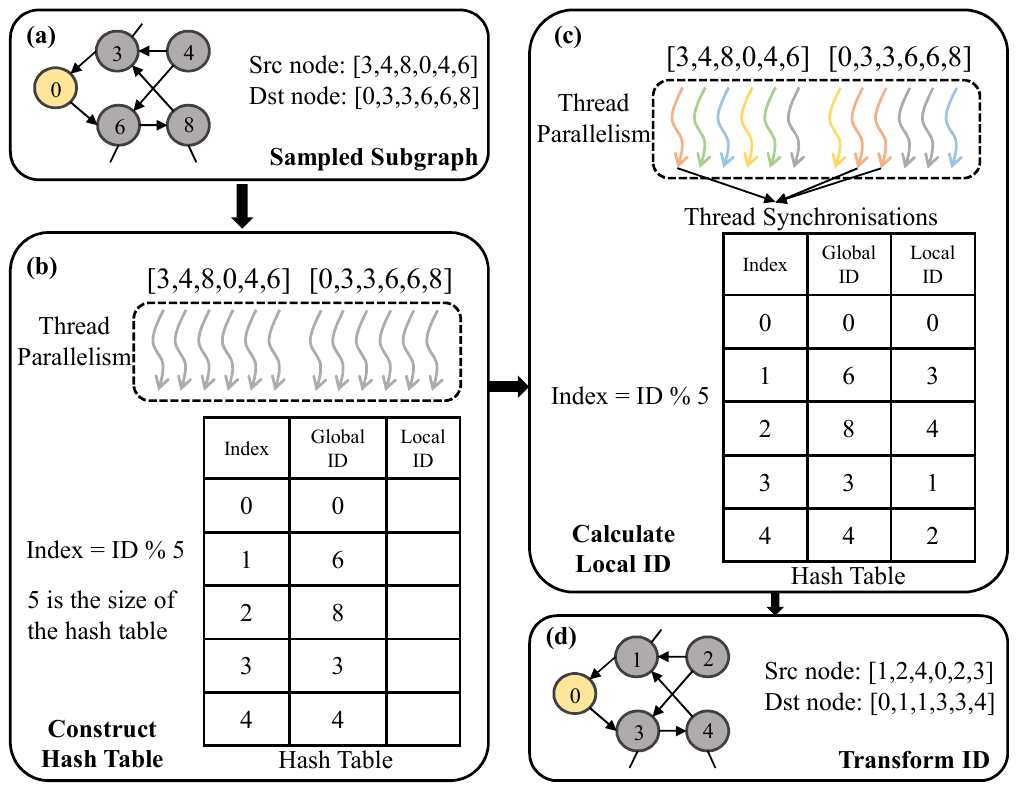}
  \caption{Illustration of the ID map process.}
\label{id_map}
\end{figure}

\subsection{Bottleneck Existing in the Sample Phase}
\label{sample_analysis}

Our breakdown analysis in Figure \colorref{time_breakdown} (`Naive+MR+MA') indicates that 
the 
sample phase now impedes the performance boosting with the two above optimizations, 
which costs more than 50\% time of the overall training process.
Through a further analysis, we observe that the ID map process takes up to 70\% 
time of the sample phase. 
As detailed in Figure \colorref{id_map}\textcolor{ACMPurple}{b}-\textcolor{ACMPurple}{d}, 
the ID map comprises 
three steps:
(1) construct a hash table for fast global ID indexing, 
(2) obtain the local ID
and record the map between global ID and local ID, and (3) transform
global IDs to local IDs using the hash table.

DGL \colorcite{wang2019deep} accelerates step (1) and (3) 
by leveraging GPU's massive threads parallelism, assigning each thread to a single global ID. 
However, the concurrent computation of local IDs in stage (2) necessitates 
synchronization to prevent multiple threads from 
repeatedly
accumulating for the same global ID, which would 
lead to conflicting mappings. 
For example, 
without 
synchronizations, the threads processing the global ID 3 will accumulate three times, 
resulting in 
the global ID 3 corresponding to multiple local IDs.
The extensive thread synchronizations considerably slow down the ID map. 

To alleviate this problem, 
we
propose the Fused-Map sampling approach to perform the ID map process in a
fused mechanism, which avoids the numerous thread synchronizations and  
expedites the sample phase (`FastGL' of Figure \colorref{time_breakdown}).

\section{Design}

In this section, we present the \textbf{FastGL} framework equipped 
with our three proposed innovative techniques, 
which
tackles the above bottlenecks well and 
substantially {unlocks} the acceleration 
potential of GNN 
training on GPU.

Figure \colorref{overview} shows the overall architecture of our FastGL framework. 
From the start of each 
epoch, the Map-Fused Sampler samples $n$ mini-batches. 
Then FastGL reorders the 
computation order of the $n$ mini-batches with the Reorder strategy. Once the reorder is done, 
the data of the
first mini-batch are loaded 
to GPU to participate in the forward and backward computations, 
which would be accelerated by our Memory-Aware computation method. 
After completing the training of the current mini-batch, FastGL loads the nodes features 
required by the next 
mini-batch through our Match process to accelerate the memory IO phase. 
We repeat the above process for the 
next $n$ mini-batches until finishing an epoch. 
$M$ GPUs complete 
the training in a data parallel manner. 

\begin{figure}[t]
  \centering
  \includegraphics[scale=0.55,trim=7.3cm 6.4cm 8cm 3.5cm,clip]{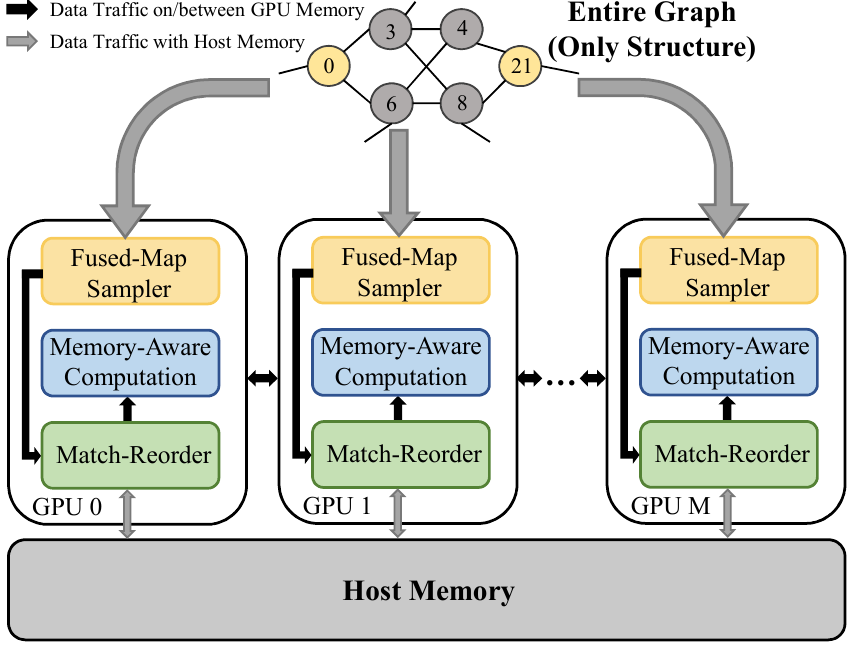}
  \caption{{Overall architecture of FastGL.}}
\label{overview}
\end{figure}

\begin{table}[t]
  \caption{The average match degree and the differences in match degrees between mini-batches on different graphs.
  $\bigtriangleup  M$ denotes the difference between the 
maximum and minimum $M_{ij}$ in one epoch.}
  \label{match_degree}
  \begin{center}
      \begin{tabular}{ccccc}
        \toprule[1.5pt]
                     Graph & Reddit & Products & MAG     &Papers-100M \\ \midrule[1pt]
                     {$Avg(M_{ij})$} & 93.2\% & 71.4\%   & 35.3\%  &38.0\%  \\
                     $\bigtriangleup  M$ & 4.9\%  & 7.0\%   & 4.2\%  &5.3\%  \\ \bottomrule[1.5pt]
        \multicolumn{5}{l}{\small * Batch size is 8000 with uniform sampling.}
      \end{tabular}
  \end{center}
  \vspace{-0.2cm}
\end{table}

\subsection{Match-Reorder Strategy}
\label{mr_section}

{Previous works \colorcite{lin2020pagraph,yang2022gnnlab}} 
incur heavy GPU memory overhead when optimizing 
the memory IO, which is infeasible for large-scale graphs.  This calls for a more efficient 
design that reduces the data traffic without requiring any extra GPU memory.

{Jangda et. al. proposed NextDoor \colorcite{jangda2021accelerating} and}
qualitatively revealed 
that the same node may appear in different sampled subgraphs.  
Our in-depth quantitative analysis further demonstrates that
there is a 
large number of overlapping nodes between different sampled subgraphs thanks to the 
complex topology of a graph. {$M_{ij}$ is the $match \ degree$ to represent the ratio of 
overlapping nodes between subgraph $i$ and subgraph $j$, i.e., 
$M_{ij} =\frac{N_o}{\min (N_i,N_j)}$, where $N_o$ is the number of 
overlapping nodes
and $N_i$/$N_j$ is the 
total number of the nodes in subgraph $i$/$j$.}
Table \colorref{match_degree} presents the average $match \ degree$ in 
one epoch on different graphs, which could be up to 93.2\%.

\begin{figure}[t]
  \centering
  \includegraphics[scale=0.5,trim=6.4cm 4.5cm 6.6cm 4.7cm,clip]{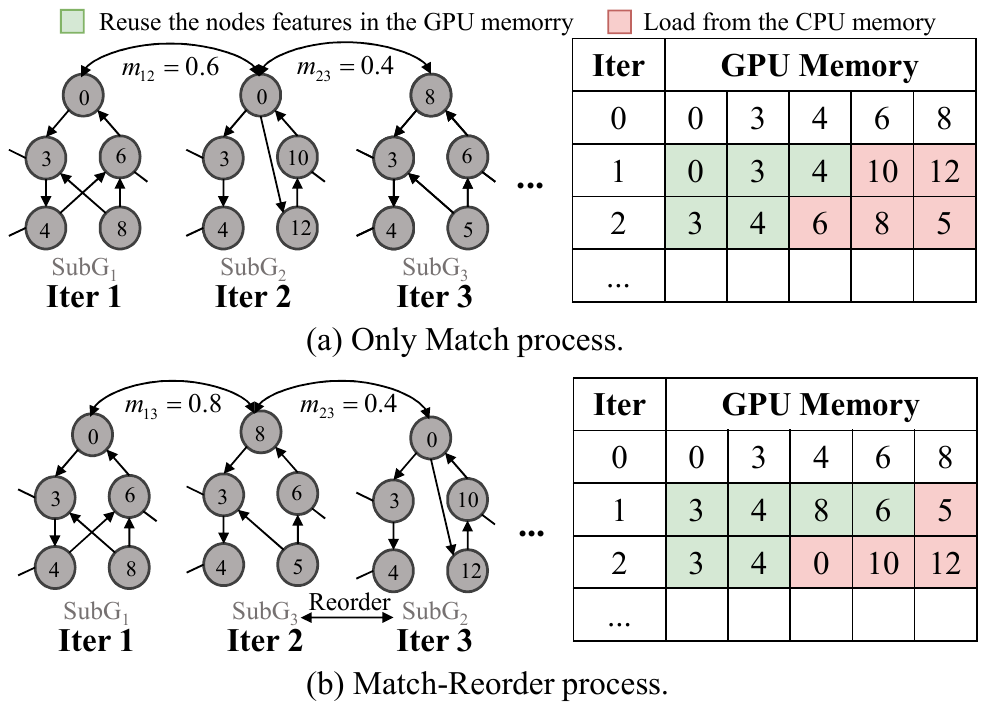}
  \vspace{-0.2cm}
  \caption{{An example of our Match-Reorder method. $m_{ij}$ denotes the 
  match degree between the subgraph $i$ and $j$.}}
  \vspace{-0.2cm}
  \label{match_reorder_example}
\end{figure}

\begin{algorithm}[t]

    \small
    \renewcommand{\algorithmicrequire}{\textbf{Initialize:}}
    \renewcommand{\algorithmicensure}{\textbf{Output:}}
    \caption{Greedy Reorder Strategy}
    \label{reorder_alg}
    \begin{algorithmic}[1]
      \STATE \textbf{Initialize:} $NodeIDList=\{NodeID_1, NodeID_2,...,NodeID_n\}$, $NodeID_i$
      is the node IDs in the $i$-th mini-batch;
      $ReorderedList$: the list to store the reordered mini-batches.
        \STATE \textbf{Greedy Reorder} ($n$):
        \STATE  Calculate the $match \ degree$ matrix $\mathbf{M}$, $m_{ij}$ is the 
        $match\ degree$ between the \textit{i-th} and the \textit{j-th} mini-batches
        \STATE  $ReorderedList$.insert($SubG_1$, 1)
        \STATE  $z = 1$ \textcolor{blue!50}{\# the index of the last inserted mini-batch}
        \FOR{$i = 2$ to $n$}
        \STATE  $h = \arg \underset{k}{\mathop{\max }}\ m_{zk}$ 
        \STATE  $ReorderedList$.insert($SubG_h$, i)
        \STATE  Set the \textit{z-th} row and column of $\mathbf{M}$ to $0$
        \STATE  $z = h$
        \ENDFOR
        \STATE  \textbf{return} $ReorderedList$
        \STATE \textbf{end}
      \end{algorithmic}
  \end{algorithm}

Therefore, 
we propose the \textbf{\textit{Match}} method, which 
performs the match before loading the nodes 
features of a new mini-batch except the first one 
to reuse the features of overlapping nodes. 
We first calculate the intersection (denoted as $OverlapNodeID$) of the nodes in the 
mini-batch
to be computed on GPU and the nodes of the 
last mini-batch. 
Then we obtain the global IDs of the nodes  
required to be loaded from the host memory to 
GPU memory ($LoadNodeID$) through subtracting $OverlapNodeID$ from the 
node sets of the mini-batch
to be computed.
GPU only loads the features of the nodes whose IDs are in the $LoadNodeID$.
The \textbf{\textit{Match}} process reduces the data
traffic by reusing and does not require 
any extra
consumption 
of GPU memory because the memory occupied by the last mini-batch is necessary.
{As an example shown 
in Figure \ref{match_reorder_example}(a), when finishing the 
training of the first subgraph ($SubG_1$) in the 1-st iteration, GPU 
only loads the features of 
node 10 and 12 from CPU
and reuses the 
features of the overlapping nodes, i.e., node 0, 3 and 4.
}

However, as the $\bigtriangleup  M$ shown in Table \colorref{match_degree}, 
the $match\ degree$ varies widely between different pairs of sampled 
subgraphs. Therefore, the computation order of the mini-batches would
affect the reuse of the overlapping nodes, which may be sub-optimal 
under the default sample order.

To mitigate this problem, we propose the \textit{\textbf{Greedy 
Reorder Strategy}} to maximize the reuse by reordering the 
computation order of the sampled mini-batches in a greedy manner. 
As shown in Algorithm \colorref{reorder_alg}, we first 
sample \textit{n} mini-batches at a time 
and initialize the \textit{NodeList} using the global IDs of the nodes in each mini-batch (line 1).
We then
calculate the $match \ degrees$
between the $n$ subgraphs (line 3) and add the \textit{1-st} sampled
subgraphs ($SubG_1$) into the \textit{ReorderedList} as the first mini-batch (line 4). 
Then, we search for the subgraph that has the maximum 
$match\ degree$ with the subgraph last added into the \textit{ReorderedList} and insert this subgraph
into the \textit{ReorderedList} (line7-8). 
To avoid inserting the same subgraph,
we set the $match \ degrees$ of
subgraphs that have been inserted to zero (line 9).
After conducting the above process for all $n$ 
mini-batches, the \textit{(i+1)-th} mini-batch
in the \textit{ReorderedList} has 
the maximum $match\ degree$ with the mini-batch at 
the \textit{i-th} location. 
{By performing the 
computation as the order in the \textit{ReorderedList} and 
alternating the mini-batches with the \textit{\textbf{Match}} process, 
we can significantly improve 
the reuse of the overlapping nodes.} 
When these $n$ mini-batches complete the training, 
we repeat the above process for the next $n$
mini-batches until finishing one epoch.

{As shown in Figure \ref{match_reorder_example}(b), 
after applying the Reorder strategy, the 
execution order of the $SubG_2$ and $SubG_3$ should be swapped because 
$m_{13}>m_{12}$
when 
fixing the order of the $SubG_1$. We can observe that our greedy 
Reorder strategy further reduces the 
memory traffic between CPU and GPU.}

\subsection{Memory-Aware Computation}

As discussed in Section \colorref{memory_aware_motivation}, the low L1/L2 cache hit rate incurred by 
irregular memory accesses during the aggregation limits the GPU performance and 
slows down the computation phase. 
As presented in Equation \colorref{aggregate_information},
the features of the source node ($x_v$), 
weight $w_{uv}$ 
and partial sums ($\hat{h_u}$) of the hidden features $h_u$ participate in the aggregation of the 
target node $u$.
In the naive case (e.g., DGL and PyG), all the above data 
is stored in the low-bandwidth global memory and is first 
fetched to high-bandwidth
L1 cache when it is requested. 
However, as shown in Figure \colorref{memory_aware}\textcolor{ACMPurple}{a},
due to the extreme sparsity and the massive data size, 
the fetched data may no longer be 
used and will be evicted to free up memory for other requested data, 
which results in the underutilization of the L1 cache and then impedes
the overall bandwidth and GPU performance.
Intuitively, storing all required data in the L1 cache would maximize the performance. 
However, two challenges exist:
(1) We cannot store the data in the L1 cache explicitly because 
the L1 cache is not software-managed. 
(2) The capacity of the L1 cache is too small (e.g., 128KB per SM in 3090 GPU) 
to accommodate all required data.

\begin{figure}[t]
  \centering
  \vspace{-0.1cm}
  \includegraphics[scale=0.53,trim=7cm 3.6cm 6.8cm 4cm,clip]{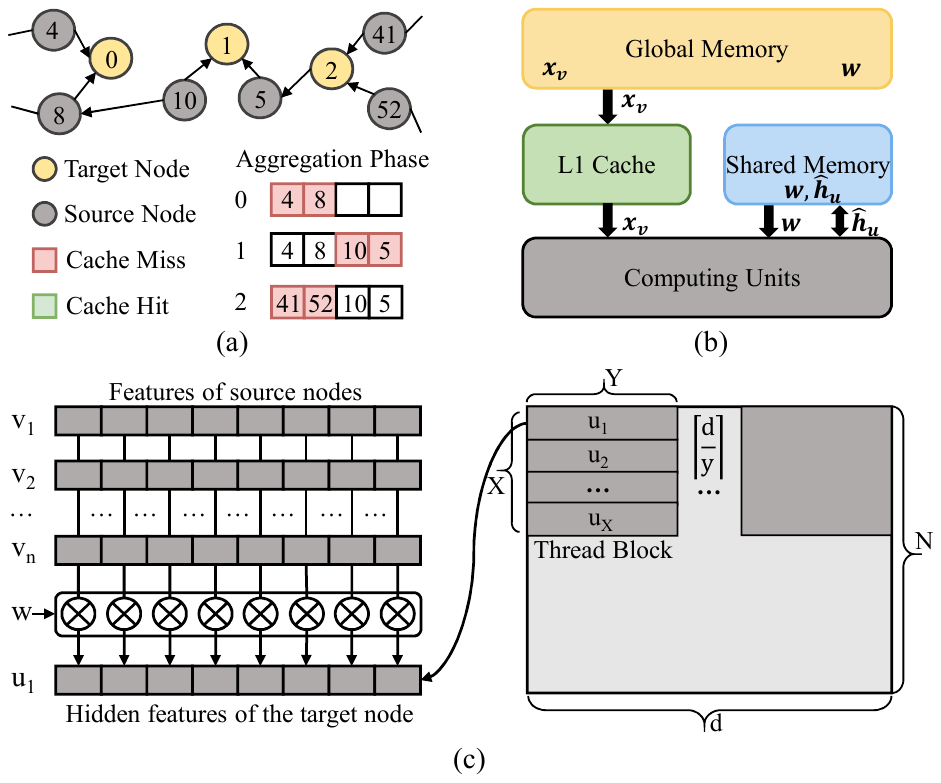}
  \caption{(a) An example of the L1 cache utilization during the aggregation 
  in the naive case. (b) The 
  memory access pattern of 
  our proposed Memory-Aware computation method. (c) The aggregation utilizing threads 
  parallelism on GPU ($N$ is the total number of target nodes). 
  $\bigotimes $ denotes the Fused Multiply Add operation.}
  \vspace{-0.3cm}
\label{memory_aware}
\end{figure}

Accordingly, we propose the \textit{\textbf{Memory-Aware}} computation method by employing 
the customizable 
shared memory provided by GPU, which 
has the same bandwidth and capacity as the L1 cache.
The key idea is that 
we adjust the memory access pattern in the aggregation phase and 
prioritize the storage of frequently
accessed data in higher-bandwidth memory
to maximally increase
the overall bandwidth
available to the computing units, thus boosting the GPU performance.

We first analyze the access frequencies of different data.
To complete the aggregation of node $u$, the features of each source node ($\bm{x}_v$) are 
read one time,
$w_{uv}$ for the node $v$ is read $d$ times where 
$d$ is the dimension of the nodes features, and 
$\bm{\hat{h}}_u$ is read $|N(u)|-1$ 
times to perform the accumulation. The access frequencies of $\bm{\hat{h}}_u$ and $w_{uv}$
significantly exceed that of $\bm{x}_v$ because the number of 
neighbors ($|N(u)|$) is often 5, 10 or 15 and 
$d$ is always large (e.g., 256 or 512).
Based on this analysis, the partial sums $\bm{\hat{h}}_u$ and 
the weight $w_{uv}$ are stored in the shared memory with higher bandwidth, and 
$\bm{x}_v$ with a lower access frequency is stored in the lower-bandwidth
global memory, as shown in Figure \colorref{memory_aware}\textcolor{ACMPurple}{b}. 
Then computing units fetch the data from its corresponding memory.
With the same lifespan as the thread block, 
the shared memory only stores the data required by the corresponding thread block, 
decreasing the memory capacity burden.
Moreover, to guarantee the correctness of the aggregation,
the aggregation of a specific target node $u$ must
be completed in one thread block.

To further illustrate the efficiency of our Memory-Aware method, 
we quantitatively analyze the memory access time as follows.
We assume that the bandwidth of the shared memory (L1 cache) and the global 
memory are $B_s$ and $B_g$, 
respectively, 
and the data is all in FP32 format.
We omit the time of accessing the L1/L2 cache for simplicity. 
In the naive case, all data should be read from the low-bandwidth global memory and 
the time of fetching data when aggregating the node $u$ is:
\begin{equation}
  \label{naive_time}
t_n = \frac{4(|N(u)|-1)d + 4|N(u)|d + 4|N(u)|d}{B_g} \ \text{.}
\end{equation}
With the Memory-Aware method, we can fetch the most accessed data, i.e., partial sums and weights, from the 
higher-bandwidth shared memory
and the access time is:
\begin{equation}
  \label{memory_aware_time}
t_m=\frac{4(|N(u)|-1)d + 4|N(u)|(d-1)}{B_s} + \frac{4|N(u)|d + 4|N(u)|}{B_g} \ \text{.}
\end{equation}
Given that $B_s\gg  B_g$, $t_m$ is notably less than $t_n$ ($\sim 10\times$).
The Memory-Aware method effectively diminishes memory access time and curtails 
the idle periods of computing units caused by delays in data retrieval,
substantially improving 
the GPU performance and accelerating the computation.

Similar to the forward pass of the aggregation, 
the target node also gathers information from its neighbors to complete the backward pass, where
the information is the  
gradient of the loss function with respect to node feature:
\begin{equation}
  \label{backward_pass}
\frac{\partial L}{\partial \bm{h}_{u}}=\sum\limits_{v\in N(u)}w_{uv}\cdot \frac{\partial L}{\partial \bm{x}_{v}} \ \text{.}
\end{equation}
Therefore, we also apply the Memory-Aware method to the backward phase
to further accelerate the computation phase.

\textbf{Thread block settings:}
{To efficiently support the fine-grained control of memory accesses,
we develop the customized CUDA kernel to perform the aggregation phase instead of adopting the traditional
GeMM kernel.} As shown in Figure \colorref{memory_aware}\textcolor{ACMPurple}{c}, we first set each thread block of the kernel
to be responsible for the concurrent aggregation of $X$ target nodes.
To improve the parallelism of the computation, 
each thread is responsible for the 
aggregation of one dimension of the hidden features. {
Because the upper limit of the threads allowed by a thread block is 
1024 in current GPU hardware, we constrain the Y-dimensional features processed by each thread 
block to suffice $X\cdot Y<1024$, 
and use $\left\lceil \frac{d}{Y}\right\rceil$ thread blocks to 
complete the aggregation of all hidden features of these $X$ target nodes. }

{When the kernel performs the aggregation,
it first fetches the corresponding features 
of source nodes from the global memory
and the weights from the shared memory. Then the kernel
conducts the multiplication to obtain 
the weighted features.  
Finally, it fetches the corresponding partial 
sums ($\hat{h_u}$) from the shared memory 
and adds the weighted features with $\hat{h_u}$
together to get the updated $\hat{h_u}$ and then writes it to the shared memory. 
}
Note that there is no requirement for thread synchronizations because each 
thread is only responsible for 
one dimension of the hidden features and
performs the
accumulation independently. 
{Each thread block only needs to store 
the $X$ Y-dimensional partial sums 
and the weights required by the $X$ target nodes in the shared memory.
The target node $u$ needs $|N(u)|$ weights.
Therefore, 
the 
capacity requirement of the shared 
memory for each thread block is $4XY+4X|N(u)|$, where $4XY$ and $4X|N(u)|$ are used to stored the 
partial sums and weights, respectively. Through setting the appropriate values of 
$X$ and $Y$, we ensure the 
size of the shared memory required by each thread block to satisfy the hardware 
limitation and keep 
the maximum occupancy of the SM.
We empirically set $X=8$ and $Y=32$.}

\begin{algorithm}[t]

    \small
    \renewcommand{\algorithmicrequire}{\textbf{Initialize:}}
    \renewcommand{\algorithmicensure}{\textbf{Output:}}
    \caption{Fused-Map Algorithm}
    \label{fused_map}
    \begin{algorithmic}[1]
      \STATE \textbf{Initialize:} 
      $HashTable$: record the mapping (global ID $\rightarrow$ local ID) 
      and enable fast indexing the mapping by global ID, {we initialize the keys of 
      the $HashTable$ to -1 and the values to 0;}
      \\ $LocalID=0$: record the current local ID.
      {\STATE \textbf{atomicCAS}($address$, $OldValue$, $NewValue$):
      \STATE \quad $ReturnVal = HashTable[HashIndex].key$
      \STATE \quad \textbf{if} $HashTable[HashIndex].key==OldValue$
      \STATE \quad \quad $HashTable[HashIndex].key==NewValue$
      \STATE \quad \textbf{end if}
      \STATE \quad \textbf{return} $ReturnVal$
      \STATE \textbf{end}}
      \STATE \textbf{InsertID}($GlobalID$):
      {\STATE \quad $InsertSucceed$ = $False$}
      \STATE \quad $HashIndex$ = \textit{HashFunction}($GlobalID$)
      {\STATE \quad \textbf{while} $InsertSucceed$ == $False$:
      \STATE \quad \quad $Val$ = $atomicCAS(HashIndex,\ -1,\ GlobalID)$
      \STATE \quad \quad \textbf{if} {$Val == GlobalID$ || $Val == -1$} \textbf{then}
      \STATE \quad \quad \quad $InsertSucceed$ = $True$
      \STATE \quad \quad \quad $Flag$ = $(Val == -1) \ ?\ True\ :\ False$
      \STATE \quad \quad \textbf{else}}
      \STATE \quad \quad \quad
      \textcolor{blue!50}{\# There is a conflict at the location of $HashIndex$}
      \STATE \quad \quad \quad
      \textcolor{blue!50}{\# The other GlobalID has been inserted at $HashIndex$}
      {\STATE \quad \quad \quad $HashIndex = HashIndex+1$}
      \STATE \quad \quad \quad
      \textcolor{blue!50}{\# Do the linear probing until the insertion successes}
      {\STATE \quad \quad \textbf{end if}}
      \STATE \quad \textbf{return} $HashIndex$, $Flag$
      \STATE \textbf{end}
      \STATE  \textbf{Fused Map}($GlobalID$):
      \STATE  \quad $HashIndex$, $Flag$ = \textbf{InsertID}($GlobalID$)
      \STATE \quad \textbf{if} {$Flag==False$} \textbf{then}
      \STATE \quad \quad $HashTable[HashIndex].value = LocalID$
      \STATE \quad \quad $atomicAdd(LocalID,1)$
      \STATE \quad \textbf{end if}
      \STATE \textbf{end}
      \end{algorithmic}
  \end{algorithm}

\subsection{Fused-Map Sampling}

As our analysis in Section \colorref{sample_analysis}, due to the extensive
thread synchronizations to obtain local IDs 
in the ID map process,
the sample phase emerges as the principal performance bottleneck 
after optimizing the memory IO and computation. 
To alleviate this problem, we propose the 
\textit{\textbf{Fused-Map}} sampling
method, which avoids the costly thread synchronizations 
by 
fusing the calculation of local IDs with 
the construction of the hash table in the ID map process.

Algorithm \colorref{fused_map} presents our Fused-Map method. 
The \textit{key} of the item in the \textit{HashTable} stores the global ID and 
the \textit{value} stores the local ID (line 1).
\textit{LocalID} records the number of unduplicated global IDs 
that have been processed, i.e., the current local ID.
To construct the hash table, the \textit{InsertID} function (line 26) inserts 
the global ID into the \textit{HashTable} as the \textit{key} of one item, whose 
location is calculated by a \textit{HashFunction}, such as the 
mod computation (line 11). 
Note that the insertion is atomic to ensure that the 
same location of the hash table can only be accessed by a single thread at a time, 
{which is
implemented by the $atomicCAS$ function 
provided by CUDA, as shown in line 13. 
The details of the 
$atomicCAS$ are shown in lines 2-8, where the operations are performed 
in one atomic transaction.
Moreover, to resolve the potential hash conflicts, we utilize 
the linear probing to find the inserted location until the insertion 
succeeds (line 20).
If there is no conflict and the global ID has not been inserted, 
we insert the global ID into 
the \textit{HashTable} (line 5) and assign the corresponding 
boolean value to the $Flag$ (line 16)
according to the returned value of the $atomicCAS$.} 
The \textit{Flag} indicates whether the item at \textit{HashIndex}  
has been assigned the same global ID or has not been inserted.
The \textit{Flag} is set to \textit{False} 
if there is no identical global ID at \textit{HashIndex}, otherwise the opposite.
If \textit{Flag} is \textit{False},
it means that 
the current inserted global ID represents a new node,
we assign the value of \textit{LocalID} to the \textit{value} of the item 
at \textit{HashIndex} (line 28), i.e.,  
construct a mapping between the global ID (\textit{key}) and the local ID (\textit{value}).
Then, the \textit{LocalID} is incremented by 1 (line 29). 
If \textit{Flag} is \textit{True},
it means that other threads have processed this global ID, 
and we do not perform any operation.
Considering that the above process is executed 
in a large number of threads concurrently, 
in order to guarantee the \textit{LocalID} to be accumulated correctly, 
we adopt the $atomicADD$
to perform the accumulation.

\begin{figure}[t]
  \vspace{-0.3cm}
  \centering
  \includegraphics[scale=0.5,trim=6.5cm 5.8cm 6.4cm 5.6cm,clip]{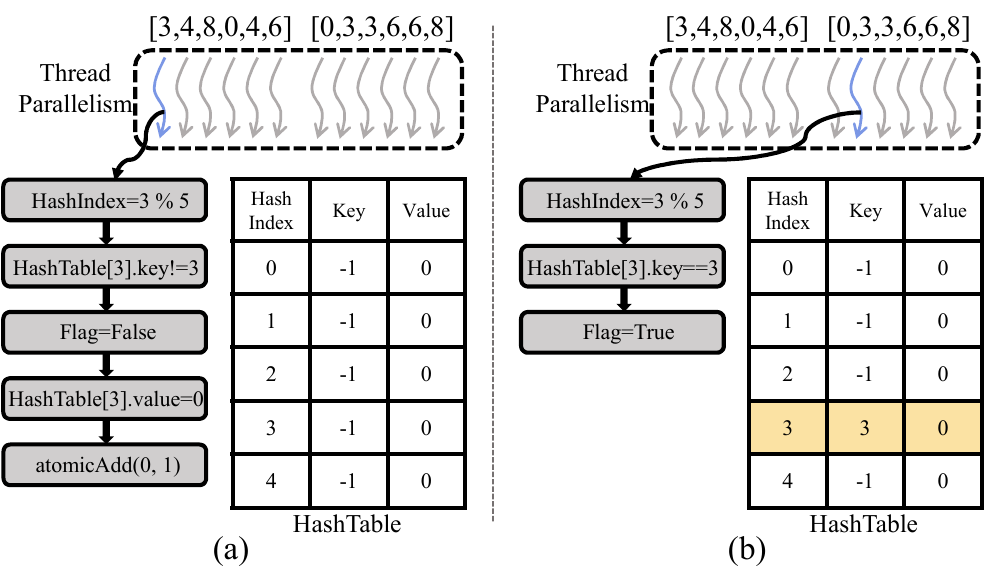}
  \vspace{-0.6cm}
  \caption{{An example of the Fused-Map process.}}
\label{fused_map_example}
\vspace{-0.5cm}
\end{figure}

{With the Fused-Map approach, we construct the hash table while obtaining the
mappings between the local IDs and the global IDs
without any thread synchronization in the same CUDA kernel, 
thus significantly accelerating the sample process.
After the construction is done, we launch another kernel and transform 
global IDs to local IDs by fast indexing 
the mapping with global ID and then complete the ID map.}

Figure \colorref{fused_map_example} illustrates an example of the Fused-Map process. 
We use $HashIndex=id\%5$ (5 is the size of the \textit{HashTable}) 
as the \textit{HashFunction} and employ the linear probing
to resolve collisions in the hash table. The initial value of \textit{LocalID} is 0.
In Figure \colorref{fused_map_example}\textcolor{ACMPurple}{a}, 
the thread atomically inserts the global ID 3 into the 
\textit{HashTable} and the global ID 3 has not been processed, thus the $Flag$ is \textit{False}.
And then we assign the current \textit{LocalID} to the \textit{value} 
of the corresponding item 
in the \textit{HashTable} and obtain the mapping between global id 3 and its local id 0.
As shown in Figure \colorref{fused_map_example}\textcolor{ACMPurple}{b}, 
when other threads try to insert global ID 3 into the table, 
this global ID has been processed, thus $Flag$ is \textit{True} and we do not perform any operation.

{\textbf{Discussion}: The users may 
be concerned about the maximum graph size 
(i.e., the number of nodes) that our Fused-Map can support 
due to the usage of the atomic functions ($atomicADD$ and $atomicCAS$). 
For the insertion 
of the pair of key and value, only the insertion of the key is atomic 
and the insertion 
of the value is determined by the returned value of 
the $atomicCAS$ (line 27-28).
And the accumulation of the $LocalID$ is performed in 
a different atomic transaction, i.e., 
$atomicADD$.
According to the 
programming guide from NVIDIA \cite{ProgrammingGuide}, 
the $atomicADD$ and $atomicCAS$ both support 64-bit 
variants. 
Accordingly, 
our Fused-Map method can support the sampling of 
graphs with up to $2^{64}$ nodes, 
which would be sufficient for the 
GNNs and graph community in the foreseeable future.}

\section{Implementation}

We implement our FastGL using the PyTorch  \colorcite{paszke2019pytorch} as the backend deep 
learning framework. We reuse the data loader of the 
open-sourced Deep Graph Library (DGL v1.0.0 \colorcite{wang2019deep}) to load the 
graph datasets and the 
sampled subgraphs. To perform our Match-Reorder strategy, we implement two Python classes, 
\textit{reorder( )} and \textit{match( )}, to perform the Greedy Reorder Strategy and 
the Match process, respectively. 
When there is sufficient GPU memory left, FastGL also 
utilizes a portion of GPU memory as a cache
to further accelerate the 
memory IO as GNNLab \colorcite{yang2022gnnlab}.
We customize CUDA kernel functions 
to achieve fine-grained control over memory access in the
Memory-Aware computation and warp these functions into user-friendly APIs, 
i.e., \textit{A3.forward()} and \textit{A3.backward()}, which are conveniently 
adopted to build layers for various GNN models. We reconstruct the sampler through the proposed 
Fused-Map sampling method in CUDA and integrate this sampler with the data loader of DGL. 
We adopt the NCCL \colorcite{nccl} 
to perform the gradient synchronization and implement the data parallel sampling-based 
GNN training on multiple GPUs in one machine. 
With the warped user-friendly APIs, 
each contribution in 
our design 
can also be conveniently applied to improve other frameworks with minimal change.

\begin{table}[t]\scriptsize
  \caption{{The configurations of the compared frameworks.}}
  \label{framework_config}
  \begin{center}
    \begin{tabular}{ccccc}
      \toprule[1.5pt]
      Frameworks & \begin{tabular}[c]{@{}c@{}}Sample\\  Device \end{tabular} & \begin{tabular}[c]{@{}c@{}}Sample\\  Optimization\end{tabular} & \begin{tabular}[c]{@{}c@{}}Memory IO\\   Optimization\end{tabular} & \begin{tabular}[c]{@{}c@{}}Computation\\   Optimization\end{tabular} \\ \midrule[1pt]
      PyG    & CPU           & No                                                                  & Prefetch                                                              & No                                                                       \\
      DGL    & GPU           & No                                                                  & Prefetch                                                              & No                                                                       \\
      {GNNAdvisor} &{No}          & {No}                                                            & {No}                                                                 & \multicolumn{1}{c}{\begin{tabular}[c]{@{}c@{}}{2D workload}\\ {management}\end{tabular}}                                                                        \\
      GNNLab & GPU           & Parallel                                                            & Cache                                                                 & No                                                                       \\
      FastGL  & GPU           & Fused-Map                                                           & Match-Reorder                                                         & Memory-Aware                                                             \\ \bottomrule[1.5pt]
      \end{tabular}
      \vspace{-0.2cm}
  \end{center}
\end{table}

\begin{figure*}[t]
  \centering
  \includegraphics[scale=0.602,trim=0.2cm 5.5cm 0.1cm 6.5cm,clip]{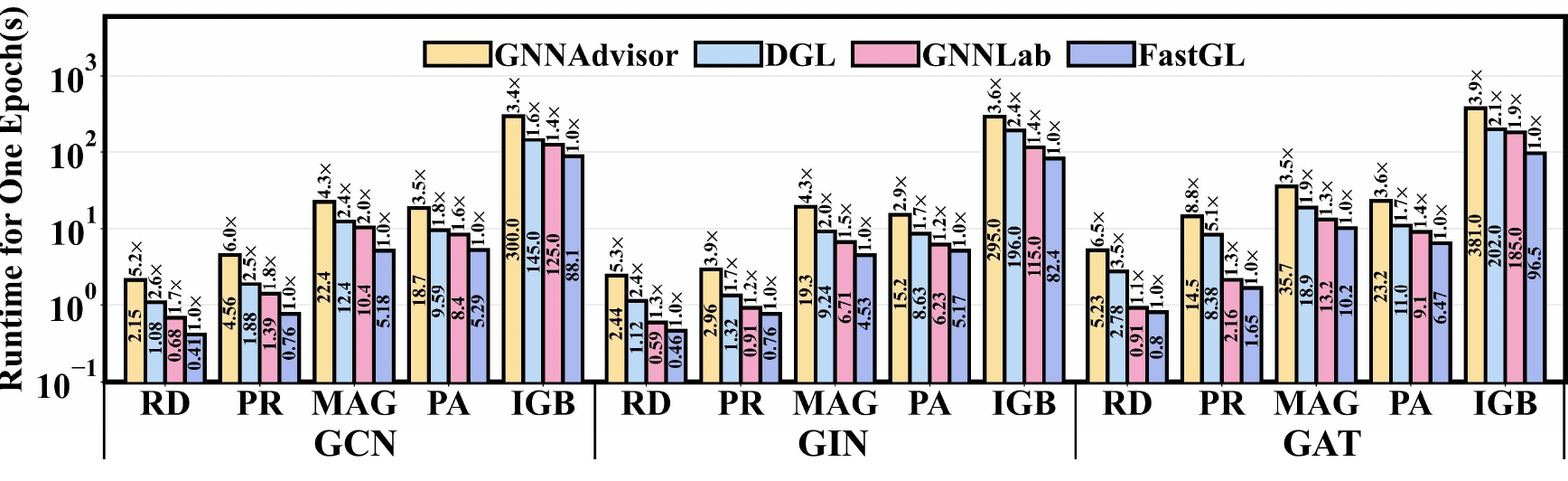}
  \vspace{-0.4cm}
  \caption{{The training speed of three GNN 
  models at various datasets on 2 GPUs compared with baselines. 
  The numbers above bars are speedups of FastGL over baselines. 
  The numbers in bars are the 
  absolute times of one epoch.}}
\label{overall_results}
\end{figure*}

\section{Evaluation}
\subsection{Experiment Setup}
\label{experiment_setup}
\textbf{Benchmarks:} We choose three representative GNN models widely used to evaluate the 
framework as previous works  \colorcite{yang2022gnnlab,liu2023bgl,lin2020pagraph}: 
Graph 
Convolutional Network (GCN) \colorcite{kipf2016semi},  Graph Isomorphism 
Network (GIN)  \colorcite{xu2018powerful} and 
Graph Attention Network (GAT)  \colorcite{velivckovic2017graph}. 
These models all adopt 3-hop
random neighborhood
sampling \colorcite{zheng2020distdgl} and the number of sampled neighbors 
for different layers are 5, 10, and 15, respectively, 
following the settings in GNNLab  \colorcite{yang2022gnnlab}. In addition, 
we set the batch size to 8000 as GNNLab \colorcite{yang2022gnnlab}. The feature 
dimension of the hidden layers of GCN and GIN is 64, and the hidden layer of GAT has eight heads and 
the dimension of each head is 8.\\
\textbf{Datasets:} We evaluate our FastGL on five real-world graph datasets with various scales 
as shown in 
Table \colorref{graph_datasets}. The Reddit (RD) is from  \colorcite{hamilton2017inductive}. The Products (PR)
and Papers100M (PA) are provided by Open
Graph Benchmark (OGB) \colorcite{hu2020open}. 
The Products is a co-purchasing network 
, and the Papers100M is 
a citation network. {Obtained from the Illinois 
Graph Benchmark (IGB) \cite{khatua2023igb},
IGB-large (IGB) graph is a collection of academic graphs.}
The MAG processed by \colorcite{bojchevski2020scaling} is a graph that contains 
scientific publication records and citation relationships between those publications.\\
\begin{table}[t]
  \caption{{The statistics of datasets used in this work (M: Million, B: Billion).}
  \label{graph_datasets}}
  \begin{center}
    \begin{tabular}{ccccc}
    \toprule[1.5pt]
    Dataset    & Nodes   & Edges & Features & Classes \\ \midrule[1pt]
      Reddit    & 232,965 & 0.11B & 602      & 41      \\
      Products   & 2.44M   & 123M  & 200      & 47      \\
      MAG        & 10.1M   & 0.3B  & 100      & 8       \\
      {IGB-large}  & {100M}   & {1.2B}  & {1024}      & {19}      \\
      Papers100M & 111M    & 1.61B & 128      & 172      \\ \bottomrule[1.5pt]
      \end{tabular}
  \end{center}
  \vspace{-0.1cm}
\end{table}
\textbf{Baselines\footnote{We use all baselines' open-sourced implementations.}:} We 
compare FastGL with the state-of-the-art PyG {2.1.0} \colorcite{lerer2019pytorch}, 
DGL {1.0.0} \colorcite{wang2019deep}, 
GNNAdvisor \cite{wang2021gnnadvisor} and 
GNNLab \colorcite{yang2022gnnlab}. 
Table \colorref{framework_config} presents the configurations of these compared frameworks.
PyG conducts sampling on the CPU. However, DGL utilizes GPU to accelerate the sample phase. 
PyG and DGL both optimize the memory IO by 
prefetching. GNNLab proposes
a more efficient cache policy to reduce data traffic and performs the sample and computation 
in parallel on different GPUs to accelerate the training.
GNNLab significantly 
outperforms other GNN frameworks \colorcite{lin2020pagraph,wang2019deep,lerer2019pytorch}. We 
make an in-depth comparison with GNNLab in
our evaluation. {Dedicated to the full-graph training, GNNAdvisor can not support sampling.
To compare against the GNNAdvisor, we 
integrate the sampler of DGL into the GNNAdvisor to enable it
to support the sampling-based GNN training.
GNNAdvisor first preprocesses
the graph according to the properties of {the GNN and graph.
Then it} accelerates the
computation by its 2D workload management. 
In the full-graph training scenario, the preprocess is just performed 
once before training, and the preprocessing time can be 
significantly amortized by the 
multiple training epochs (e.g., 200). Therefore, the preprocessing 
time can be overlooked. 
However, for the sampling-based training, 
only after the sampled subgraphs are preprocessed, 
can the computation be performed in each iteration, i.e., each iteration needs 
preprocessing. Thus the 
preprocessing time must be included in the overall end-to-end training time 
to ensure a fair comparison with other sampling-based training systems.
}
All results are the average values over 20 epochs.\\
\textbf{Environment:} All evaluations are performed on 
a GPU server that consists of two AMD EPYC 7532
CPUs (total $2\times 32$ cores), 512GB DRAM, and eight NVIDIA
GeForce RTX 3090 (with 24GB memory) GPUs. The software 
environment of the server is configured with Python v3.7.13,
PyTorch v1.10.1, CUDA v11.0, DGL v1.0.0, and PyG v2.1.0.

\vspace{-0.4cm}

\subsection{Overall Performance}

As shown in Figure \colorref{overall_results}, we compare the training speed with baselines on 
three GNN models over various datasets. The time is measured on 2 GPUs because GNNLab has to
run on at least 2GPUs to obtain optimal performance where one GPU is to sample and the other one to perform the computation.

By accelerating the memory IO, sample, and computation phases simultaneously, 
FastGL significantly outperforms 
PyG, DGL, {GNNAdvisor} and GNNLab 
by up to $28.9\times$ (from $4.3\times$), $5.1\times$ (from $1.7\times$),
{$8.8\times$ (from $2.9\times$)} and $2.0\times$ 
(from $1.1\times$), 
respectively. {We do not plot the results of PyG in Figure \ref{overall_results} because
PyG is more than an order of magnitude slower than our FastGL.}
The speedup over PyG is
especially significant because PyG performs the sampling on the CPU, which is 
ill-suited to large-scale graph scenarios that require massive parallel operations,
resulting in the sample phase occupying up to 97\% of the overall time.
Our FastGL optimizes the sampling by the Fused-Map method, 
which fully takes advantage of the thread parallelism on GPU and considerably 
reduces the sampling time. 
DGL also accelerates the sample utilizing GPU, but the vast volume of 
thread synchronizations still hinder the acceleration of sampling, 
which is tackled well by our FastGL.
Compared with GNNLab, FastGL also obtains substantial speedups. Although GNNLab 
optimizes the memory IO phase, the benefits discount on large-scale graphs where there is no 
extra GPU memory to be used as a cache (e.g., MAG and PA). However, on these datasets, 
FastGL also achieves considerable speedups thanks to the Match-Reorder strategy, 
which significantly 
reduces the data traffic without any extra GPU memory.
{Although GNNAdvisor optimizes the computation phase, the necessary
preprocess in each iteration incurs significant time overhead, 
which leads to GNNAdvisor performing poorly in the sampling-based training scenario.}
However, by utilizing our
Memory-Aware method without any preprocessing, 
FastGL accelerates the computation 
and further boosts the overall training process.
Additionally, FastGL consistently obtains significant speedups on all three GNN models, which demonstrates 
the generalization of our FastGL. 

\begin{figure}[t]
  \centering
  \vspace{-0.3cm}
  \subfloat[]{\includegraphics[scale=0.3]{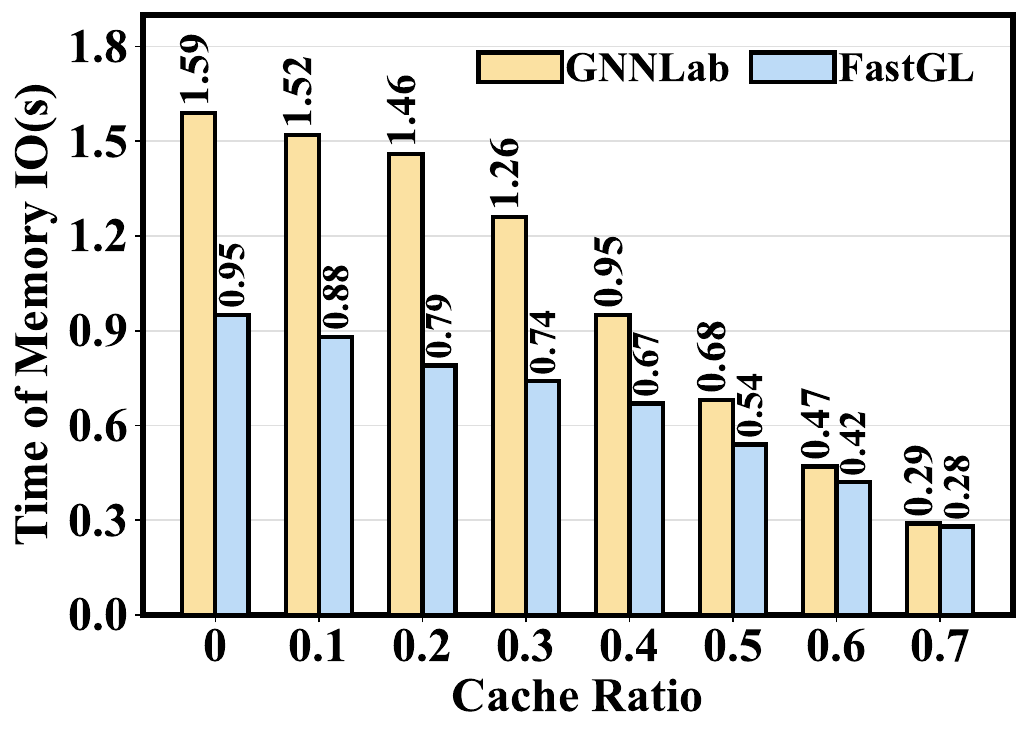}\label{match_cmp}} 
  \subfloat[]{\includegraphics[scale=0.3]{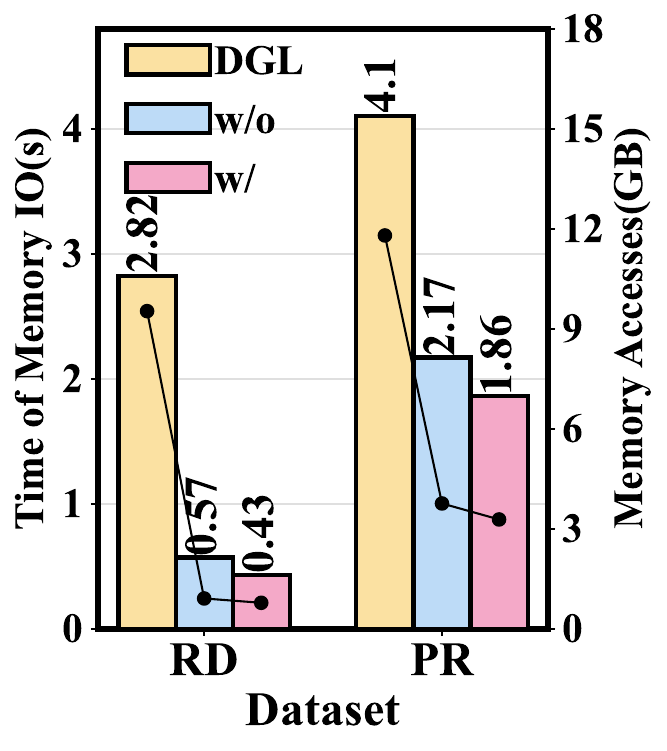}\label{reorder_cmp}} 
  \vspace{-0.2cm}
  \caption{The time spent on the memory IO comparison between (a)
  GNNLab and FastGL of GCN on Products; 
  {(b) with and without the Greedy Reorder Strategy on GCN 
  (measured on one GPU).}}
  \vspace{-0.2cm}
\label{match_reorder_exp}
\end{figure}

\subsection{Performance Breakdown}
To analyze where our gains come from and demonstrate 
the effectiveness of our three proposed techniques separately,
we conduct more detailed breakdown analyses.

\begin{table}[t]
  \caption{The comparison of time spent 
  on memory IO between DGL and FastGL on GCN on 1 GPU when 
  employing the random walk sampler.{`FastGL-nG' denotes 
  the situation without our Greedy Reorder Strategy.} 
  Values in ( )
  are normalized speedup ratios.}
  \label{other_sampling_method}
  \begin{center}
    \begin{tabular}{ccccc}
      \toprule[1.5pt]
             & RD   & PR   & MAG  & PA   \\ \midrule[1pt]
      DGL    & 1.91s (1.0) & 1.92s (1.0) & 16.4s (1.0) & 4.18s (1.0) \\
      {\begin{tabular}[c]{@{}c@{}}FastGL-\\  nG \end{tabular}}  & {0.75s (2.6)} & {1.30s (1.5)} & {15.1s (1.1)} & {3.89s (1.1)} \\
      FastGL & 0.65s (2.9) & 1.16s (1.7) & 13.1s (1.3) & 3.54s (1.2)   \\ \bottomrule[1.5pt]
      \end{tabular}
  \end{center}
\end{table}

\textbf{1) The effectiveness of the Match-Reorder strategy:} The Match-Reorder is proposed to reduce 
the data traffic without imposing any GPU memory cost.
We compare the time of memory IO with GNNLab under different 
memory capacity remaining situations, as shown 
in Figure \colorref{match_cmp}.
We cache a portion of nodes features into GPU 
memory (referred to as $cache\  ratio$) to simulate how much GPU 
memory is left. For example, \textit{cache ratio = 0} represents that 
there is no GPU memory to be used as the cache, and  $cache \ ratio = 0.1$ represents that
there is $0.1\cdot M$ memory left on GPU, where $M$ is the total size of the nodes features.
In large-graph scenarios, there 
is often a little memory left to be used as the cache (\textit{cache ratio < 0.5}) and  
Figure \colorref{match_cmp} reveals that 
our 
Match-Reorder approach significantly reduces the time of the memory IO in these situations, 
which demonstrates the 
superiority of our Match-Reorder strategy. 
Additionally, when sufficient GPU memory is available for caching, 
our method also has a minor improvement over GNNLab. 
We also explore the impact of the Greedy Reorder Strategy on the memory IO time. 
As shown in Figure \colorref{reorder_cmp}, 
without the Greedy Reorder Strategy,
FastGL still obtains significant speedups compared to DGL (`w/o' v.s. `DGL') 
thanks to our proposed Match process. {The solid lines represent 
the average number of 
memory accesses per epoch.}
Moreover, equipped with the Greedy Reorder Strategy (`w/'), FastGL 
accelerates the memory IO phase by up to 25\% compared with 
the situation only with the Match process (`w/o') because 
our Greedy Reorder Strategy maximizes
the reuse of the overlapping nodes between different mini-batches and further reduces the 
data traffic based on the Match process.

\begin{figure}[t]
  \centering 
  \begin{minipage}[t]{1\linewidth}
    \centering
    \includegraphics[scale=0.408,trim=0cm 0cm 0cm 0cm,clip]{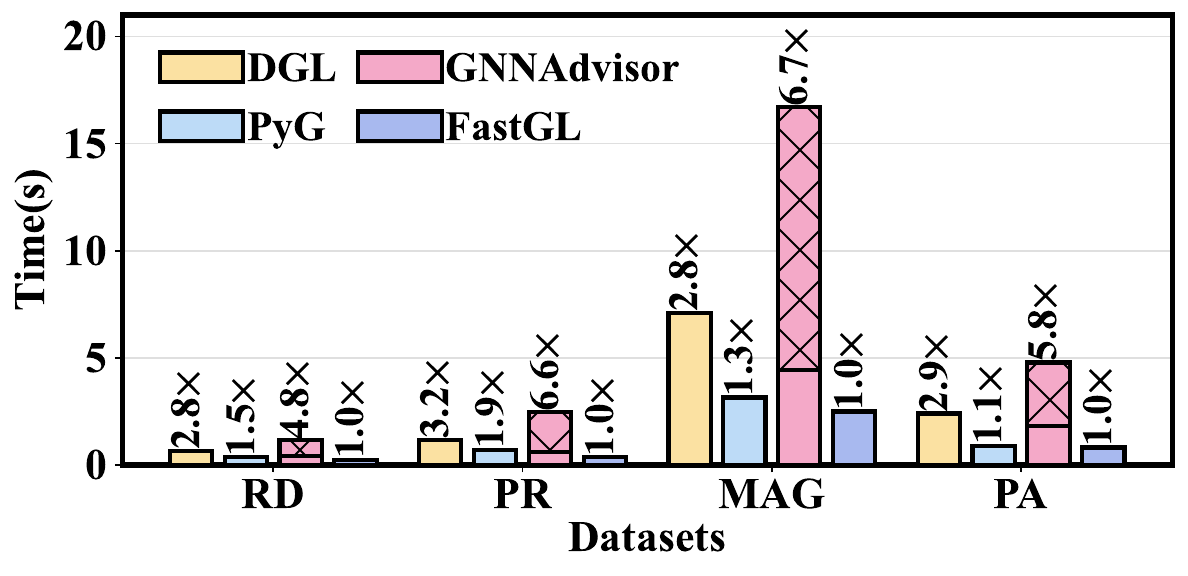}
    \vspace{-0.2cm}
    \caption{The time spent on the computation phase comparison on GCN with 2 GPUs.}
    \label{memory_aware_benefit}
  \end{minipage} 
  \\
    \begin{minipage}[t]{1\linewidth}
      \centering
      \includegraphics[scale=0.408,trim=0cm 0cm 0cm 0cm,clip]{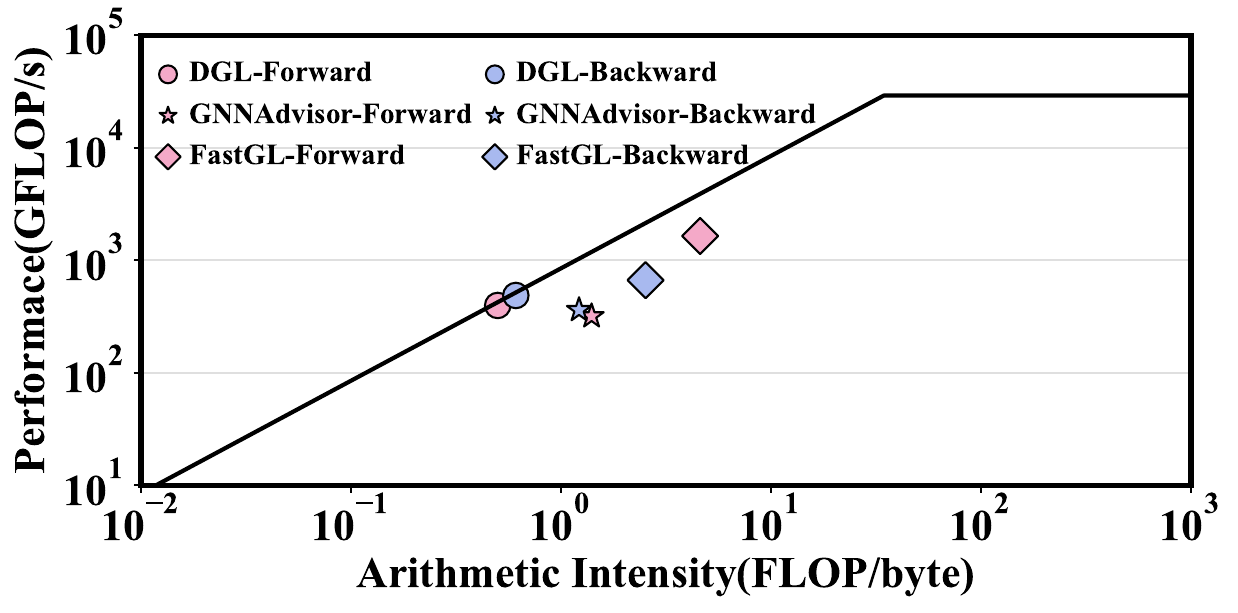}
      \vspace{-0.6cm}
      \caption{Roofline analyses of the computation phase of GCN in different frameworks on Products dataset.}
      \label{roofline}
      \vspace{-0.2cm}
  \end{minipage}
\end{figure}

The efficiency of our Match-Reorder strategy is highly related to the match degrees 
between different subgraphs, which are determined by the sampling algorithm. 
To illustrate the broad applicability 
of our approach, we compare the time spent on memory IO phase 
of FastGL and DGL on GCN when employing the random walk algorithm to sample. 
We set the 
hyperparameters of the random walk sampler (length is 3) as the 
PinSAGE \cite{ying2018graph} used, which is 
a GNN for web-scale recommender systems. As shown in 
Table \ref{other_sampling_method}, our FastGL can still accelerate the 
memory IO phase, which demonstrates the 
broad applicability of our Match-Reorder method. 
{Moreover, the comparison between
`FastGL-nG' and `FastGL' shows that our Greedy Reorder Strategy can also be applied to other 
sampling algorithms.}

\begin{figure}[t]
  \begin{minipage}[]{0.22\textwidth}
  \vspace{-0cm}
  \centering
  \includegraphics[scale=0.3,trim=0cm 0cm 0cm 0cm,clip]{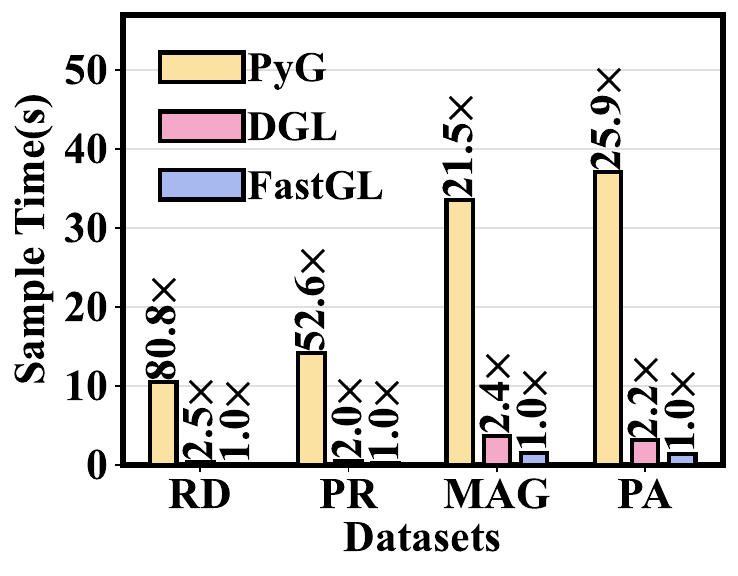}
  \vspace{-0.2cm}
  \caption{Sample time of each epoch comparison on GCN.}
  \label{sample_cmp}
  \end{minipage}
  \qquad
  \
  \begin{minipage}[]{0.2\textwidth}
    \vspace{-0.4cm}
    \captionof{table}{The spent time (s) comparison between DGL and our 
    Fused-Map method during the ID map process.}
    \label{id_map_time_cmp}
    \scriptsize
    \begin{tabular}{p{0.4cm}<{\centering}p{0.7cm}<{\centering}p{1.2cm}<{\centering}}
       \hline \toprule[1.5pt]
       & DGL  & Fused-Map \\ \midrule[1pt]
     RD  & 0.18($2.3\times$) & 0.08($1\times$)      \\
     PR  & 0.30($2.1\times$) & 0.14($1\times$)      \\
     MAG & 2.55($2.6\times$) & 0.98($1\times$)      \\
     PA  & 2.18($2.7\times$) & 0.81($1\times$)      \\ \bottomrule[1.5pt] 
    \end{tabular}
 \end{minipage}
  \vspace{-0.2cm}
\end{figure}

\begin{figure*}
  \centering
  \vspace{-0.2cm}
  \subfloat[]{\includegraphics[scale=0.23]{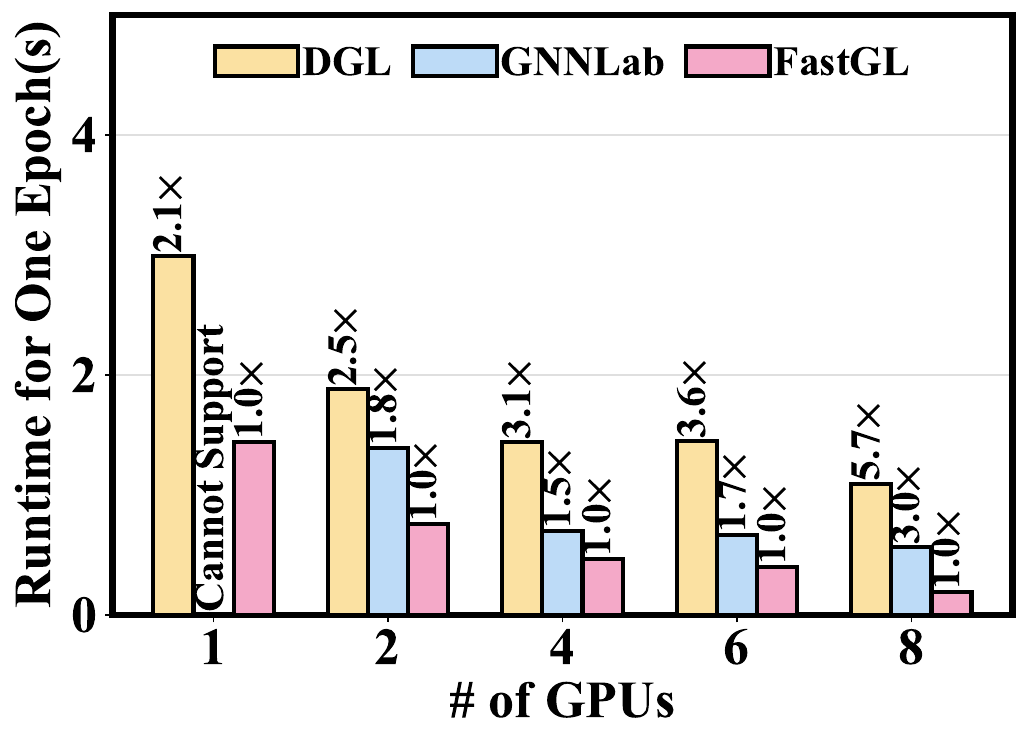}\label{scale_gpu}} \hspace{1pt}
  \subfloat[]{\includegraphics[scale=0.23]{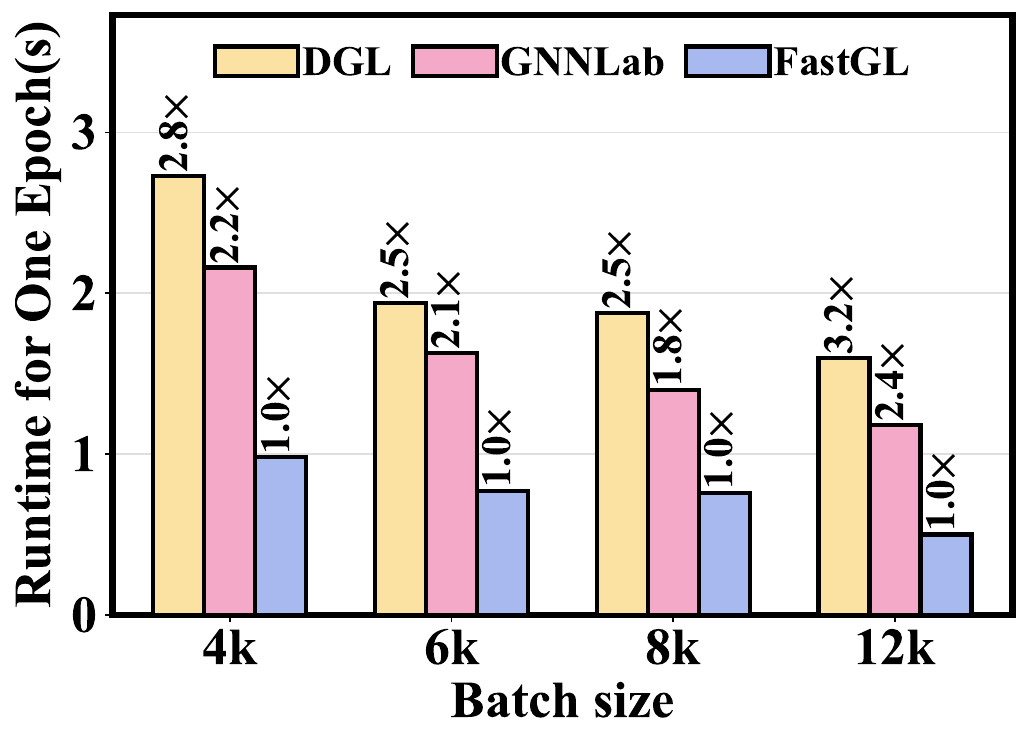}\label{scale_bs}}  \hspace{1pt}
  \subfloat[]{\includegraphics[scale=0.23]{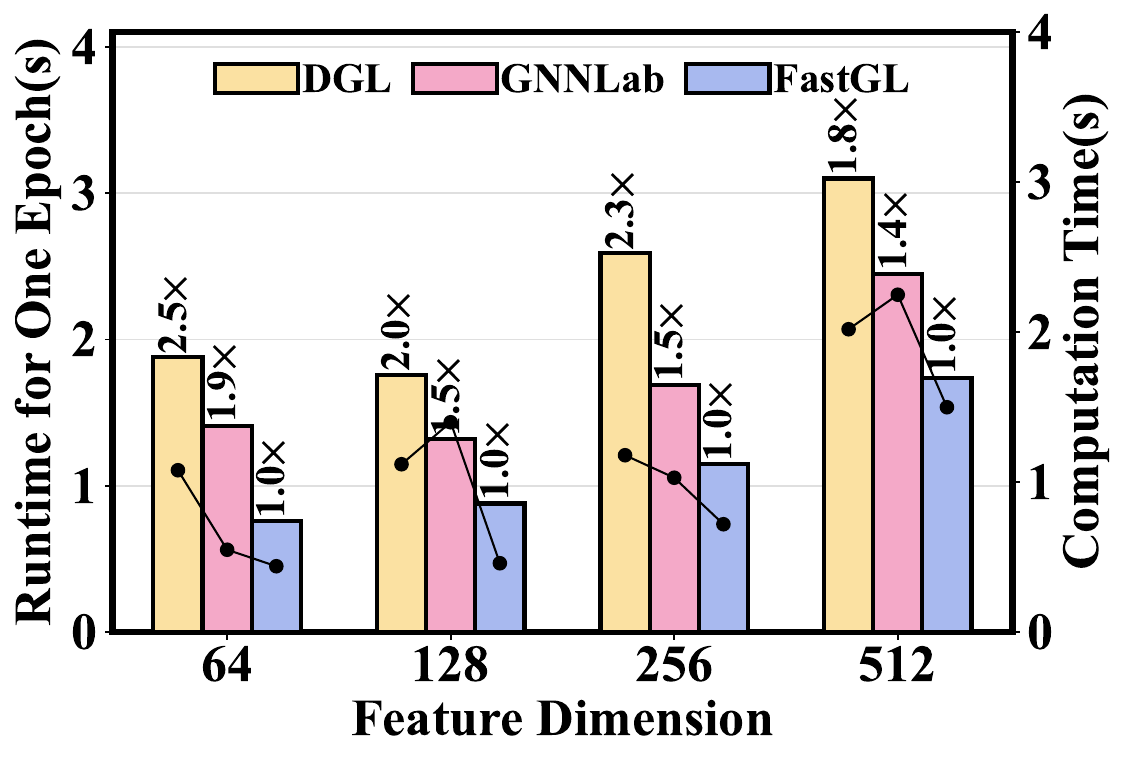}\label{scale_dim}} \hspace{1pt}
  \subfloat[]{\includegraphics[scale=0.23]{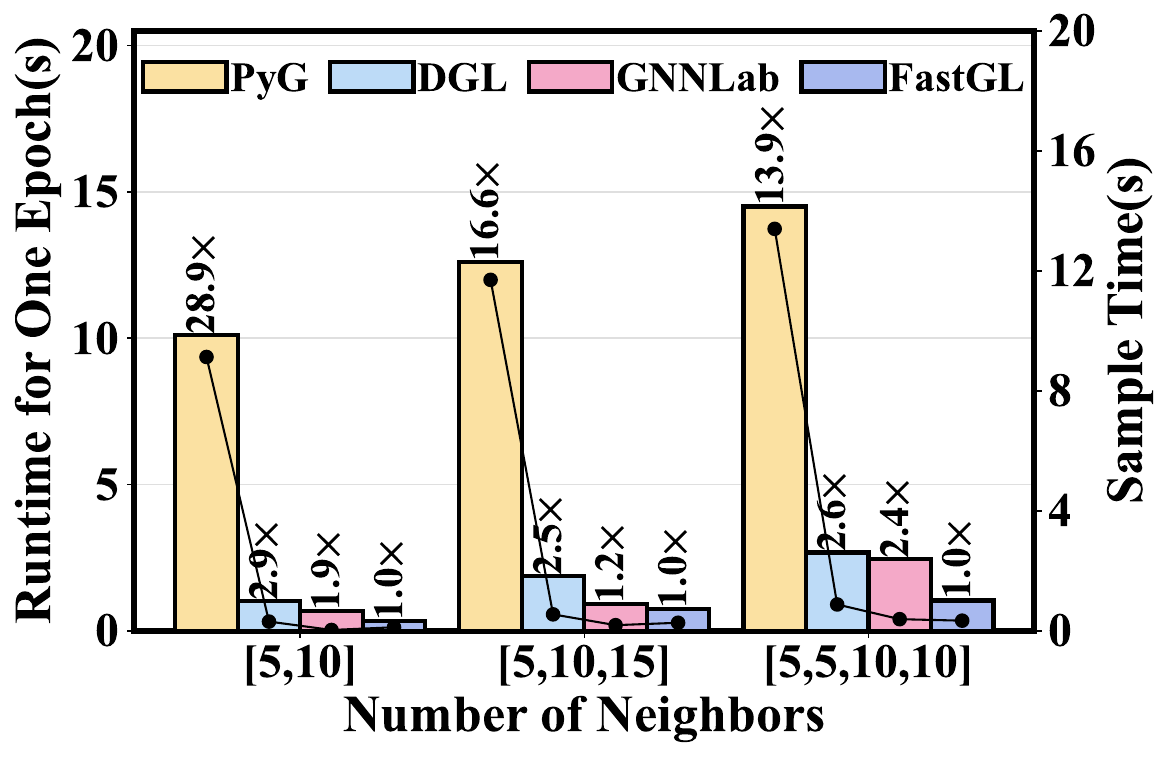}\label{scale_hop}}
  \caption{{(a) Scalability on the number of GPUs.} (b) Scalability on the batch size.
  (c) Scalability on the feature dimension. (d) Scalability on the number of sampled neighbors and the layers 
  ($[N_1,N_2,\ldots,N_k]$represents there are $k$ layers and the sampled neighbors for the 
  \textit{k-th} is $N_k$).} 
  \label{scale_results} 
\end{figure*} 

\begin{figure}[t]
  \centering
  \vspace{-0.2cm}
  \includegraphics[scale=0.6,trim=9.5cm 6.8cm 7.5cm 7.5cm,clip]{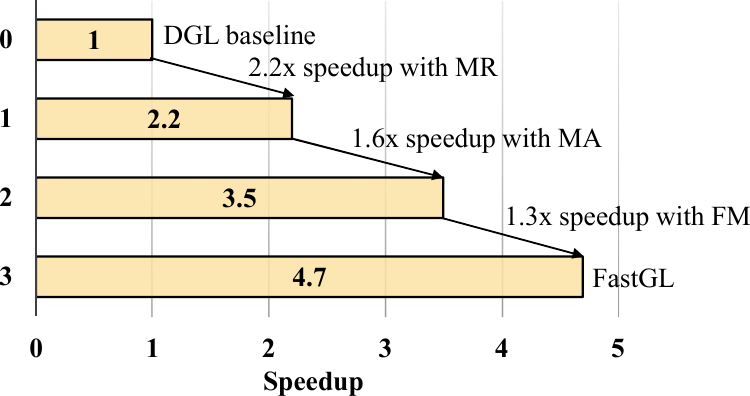}
  \vspace{-0.2cm}
  \caption{The breakdown analysis on the average overall 
  speedup across all five datasets {on GCN using 2 GPUs}. 
  `MR' denotes Match-Reorder,
  `MA' denotes Memory-Aware, and `FM' denotes Fused-Map.
  }
  \vspace{-0.15cm}
  \label{ablation_study}
\end{figure}

\textbf{2) The superiority of the Memory-Aware computation:} As shown 
in Figure \colorref{memory_aware_benefit}, we compare the computation time with DGL, PyG, and GNNAdvisor.
We can observe that our Memory-Aware method significantly outperforms the three frameworks 
on all graph datasets with $1.1\times$ to $6.7\times$ speedup in the computation phase. GNNAdvisor has a 
negative effect on accelerating the computation phase because it is dedicated to the full-batch training
and requires preprocessing the graph. 
In the full-batch training scenario, the 
preprocess only 
performs once 
and the overhead can be amortized by multiple training epochs (e.g., 200). 
However, for the large-scale 
graph where sampling-based training is necessary, 
the preprocess is performed for each sampled subgraph, and thus 
the overhead of preprocessing
cannot be amortized. {The shadowed top part of the GNNAdvisor bar in 
Figure \ref{memory_aware_benefit} displays the preprocessing
time, which occupies up to 75\% of the overall computation process and 
results in the significant deceleration of 
the computation phase.}
Furthermore, to demonstrate the effectiveness of improving the GPU performance, 
we also perform the roofline analyses for the forward and backward pass 
of the aggregation phase,
as shown in Figure \colorref{roofline}. 
Our FastGL achieves up to $4.2\times$ 
higher actual performance
compared with DGL and GNNAdvisor. 

Note that our Memory-Aware method 
do not change the 
total accessed size of data from GPU memory to computing units 
during the computation phase. However, through improving the 
achievable bandwidth by storing the 
more frequently accessed data in the 
memory with a higer bandwidth, 
the memory access time of the computation 
phase is significantly reduced, which 
accelerates the computation.

\textbf{3) The benefits from the Fused-Map sample:}
The sample phase time comparison with DGL and PyG on GCN (running on 2 GPUs)
is shown in Figure \colorref{sample_cmp}.
The sampling speed in FastGL is much faster than PyG (up to $80.8\times$ speedup). 
Compared with DGL, 
our FastGL can also obtain up to $2.5\times$ (from $2.0\times$) speedup because 
the Fused-Map sample avoids the vast volume of thread synchronizations 
to calculate the local IDs, which is the main bottleneck in the sampling by GPU. 
Moreover, 
Table \colorref{id_map_time_cmp} indicates that our Fused-Map sample method significantly 
reduce the time spent on the ID map process by the fused mechanism.

{\textbf{4) The ablation study on where the performance gains from:} 
We conduct a breakdown analysis on 
the overall speedup to analyze 
how our three proposed techniques separately affect the overall speedup.
As shown in Figure 
\ref{ablation_study}, we run 
this experiment on GCN using the five datasets and adopting DGL as 
the baseline. We can observe that optimizing the memory IO phase by our 
Match-Reorder method significantly reduces the overall training time because the 
memory IO phase dominates the training process, as we analyze in Section 
\ref{rethink_memory}. Our Memory-Aware computation manner also brings 
$1.6\times$ speedup gain by efficiently improving the achievable bandwidth of the 
computing units. The gains from the Fuse-Map method are relatively 
low compared to the above 
two, because the time spent on sampling is a small 
fraction (31\%-51\%) of the overall training time.  
}

\subsection{Scalability}
To demonstrate the generalization of our FastGL, we conduct the scalability experiments using different 
settings on GPU and GNN models. {If not specified, the GNNs 
used to evaluate are all set as described in 
Section \ref{experiment_setup}, with a batch 
size of 8000 trained on Products with 2 GPUs.}

\textbf{1) The scalability on the number of GPUs:} Figure \colorref{scale_gpu} displays 
the experimental results using different
numbers ($n$) of GPUs. {When running on 1 GPU, we only evaluate on DGL and 
FastGL because GNNLab cannot support the training on 1 GPU.} 
For GNNLab, we utilize 1 GPU to sample when $n<=4$ (the others to compute) 
and 2 GPUs to sample when $n>4$ to obtain the optimal
performance of GNNLab. 
In general, FastGL consistently outperforms other frameworks when the GPU number increases.
Moreover, as the number of GPUs increases, the speedups over 
baselines are more significant, 
which shows better scalability of FastGL. 
{Across all datasets, when using 8 GPUs,
DGL and FastGL on-average obtain $3.36\times$ and $5.93\times$ 
{speedups over that on 1 GPU}, 
respectively.}

\textbf{2) The scalability on the batch size:} We also explore how the batch size affects the training 
speed. As shown in Figure \colorref{scale_bs}, FastGL obtains $1.8\times$ to $3.2\times$ speedup with
different batch sizes 
compared to baselines. We can observe that FastGL achieves higher speedup with larger batch sizes 
(e.g., 12k v.s. 6k). The reasons are twofold: (1) When batch size is larger, there are more overlapping nodes between 
different mini-batches and our Match-Reorder strategy can reduce more data traffic. (2) 
As the batch size increases, the sample phase becomes the main bottleneck 
, which is significantly accelerated by our Fused-Map sample method.

\textbf{3) The scalability on the feature dimension:} Figure \colorref{scale_dim} presents the effect of 
feature dimension on the training speed (bar) and the computation speed (solid line) 
using different frameworks. The training speed of FastGL consistently
outperforms the other frameworks ($1.4\times$ to $2.5\times$) 
when adopting different feature dimensions. The experimental results on 
computation speed demonstrate that our Memory-Aware method is effective for various 
feature dimensions. 

\textbf{4) The scalability on the number of sampled neighbors and the hops:} We also 
investigate the impact of 
sampled neighbors and the layers on the overall time (bar) and 
the sample time (solid line) in Figure \colorref{scale_hop}. With different numbers of layers and sampled 
neighbors, FastGL obtains $1.2\times$ to $28.0\times$ training speedups over baselines, 
which demonstrates 
the robustness of FastGL. Moreover, the speed of sampling using FastGL is significantly higher than 
PyG and DGL, thanks to the fused ID map process. The sampling speed of 
FastGL is comparable to that of GNNLab on `[5,10]' and `[5,10,15]' because 
GNNLab runs the sample and computation 
in parallel on different GPUs where the latency of sampling is hidden by the computation. 
However, 
when the size of the sampled subgraph increases (`[5,5,10,10]'), 
the latency cannot be hidden 
by the computation well, where our FastGL spends significantly less time sampling than GNNLab.

\small
\begin{table}[t]
  \caption{{The comparison of the GPU memory usage on GCN 
  over various datasets on 1 GPU.}}
  \label{memory_trade_off}
  \begin{center}
    \setlength{\tabcolsep}{1.5mm}{
    \begin{tabular}{cccccc}
      \toprule[1.5pt]
             & RD     & PR     & MAG     & {IGB}      & PA        \\ \midrule[1pt]
      DGL    & 6257MB & 9853MB & 20085MB & {23447MB}  & 14030MB   \\
      FastGL & 5217MB & 7705MB & 19893MB & {21035MB}  & 14629MB   \\ \bottomrule[1.5pt]
      \end{tabular}
    }
  \end{center}
  \vspace{-0.2cm}
\end{table}

\normalsize

\subsection{Memory Trade-Off Analysis}
To analyze the memory overhead of our FastGL,
we compare the GPU memory usage 
of the GCN model at various datasets on 1 GPU with DGL.
{As shown in Table \ref{memory_trade_off}}, the memory usage of the 
two systems is comparable, which demonstrates that the memory overhead
induced by our methods is not significant. The reasons are twofold:
(1) The metadata used in our Fused-Map is also used in the DGL system,
e.g., the hash table. (2) In our Reorder strategy, we only store the 
topology information of the currently processed subgraph on the GPU 
and the others in the host memory, {which would not increase the GPU memory 
usage compared with DGL.} Note that 
we prefetch the topology information 
of the next subgraph, whose time can be perfectly overlapped 
by the computation
because its size 
(without nodes features) is minimal.

\subsection{Training Convergence}

To confirm the correctness of our FastGL, we evaluate the training loss of training
two GNN models (i.e., GCN and GIN) with FastGL and DGL (baseline) over
Reddit on 2 GPUs. As shown in Figure \colorref{train_loss}, FastGL converges
to approximately the same loss as the original DGL
when completing the overall training process (training one mini-batch is an iteration).

\begin{figure}
  \centering
  \vspace{-0.2cm}
  \subfloat[Training loss on GCN.]{\includegraphics[scale=0.28]{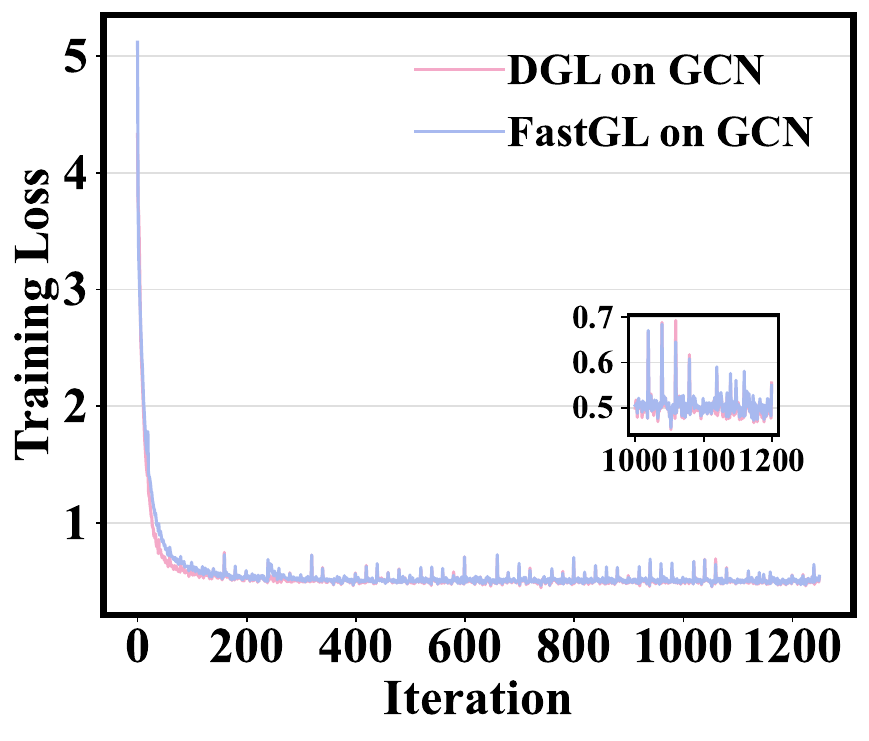}\label{loss_gcn}}
  \subfloat[Training loss on GIN.]{\includegraphics[scale=0.28]{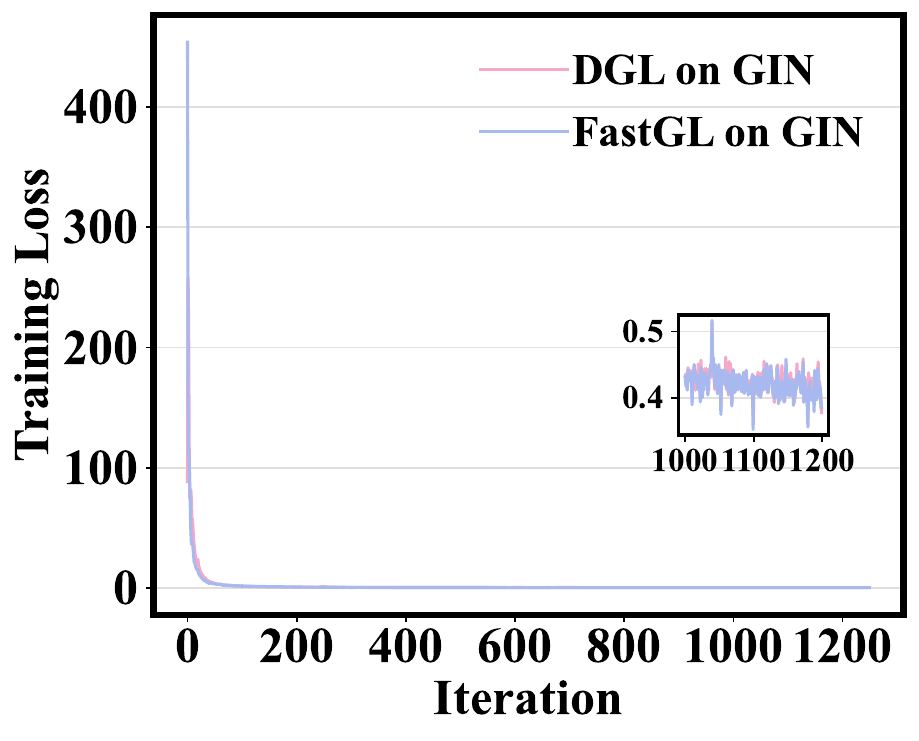}\label{loss_gin}}  
  \caption{The training loss of FastGL and DGL on Reddit dataset of GCN and GIN models.} 
  \label{train_loss} 
  \vspace{-0.4cm}
\end{figure}

\section{Discussion}
\textbf{1)The broader applicability of our FastGL:}
In this work, we focus on the acceleration of 
sampling-based GNN training on a single machine with multiple GPUs. 
Nonetheless, we expect 
that FastGL is also efficient on multiple machines 
because the mechanism of our
three contributions proposed in this paper are irrelevant 
of the number of machines. Moreover, besides the 
random neighborhood sampling algorithm mainly used in our evaluation, our
Fused-Map method can also be employed to 
accelerate diverse sampling algorithms since they all need
to transform the global ID to the local ID (ID map process) to update the subgraph structures
\colorcite{hamilton2017inductive,chen2018stochastic,chen2018fastgcn,
zou2019layer,huang2018adaptive,zhang2021biased,perozzi2014deepwalk}. 

\textbf{2)The difference between our Match-Reorder and 
NextDoor\colorcite{jangda2021accelerating}:} {NextDoor and our 
Match-Reorder method} both are based on the observation that the 
same node may appear in different sampled subgraphs. However, 
we develop completely different methods. NextDoor assigns the 
overlapping node to the same grid block to realize coalesced global 
memory accesses and accelerate the sampling phase. Differently, 
we first develop the Match strategy to reuse the overlapping 
nodes between subgraphs to reduce the data traffic between CPU 
and GPU and accelerate the memory IO phase. Then, the greedy Reorder strategy 
is proposed to maximize this reduction.

\textbf{3)The future direction of accelerating the 
sampling-based GNN training:} Our work reveals that the memory IO phase 
dominates the overall sampling-based GNN training process now.
The memory IO phase consists of  two stages: 
(1) organize the data on the CPU side to ensure they are 
consecutive in the memory; (2) transfer the data from CPU to GPU.
A fact that 
cannot be ignored is the increasing bandwidth 
between the GPU and the host, e.g., 
the upcoming widely used next-generation architecture, 
NVidia's Grace Hopper, holds up to 900GB/s of bandwidth \cite{hopper}.
In this situation, the time spent on stage (2) may not be significant, and 
the bottleneck would shift toward stage (1).
Therefore, optimizing the way data is 
organized on the CPU side may be able to lead to 
new performance breakthroughs in the future.
\endgroup

\begingroup
\spaceskip=1\fontdimen2\font plus 0.5\fontdimen3\font minus 1.8\fontdimen4{0.35cm}
\section{Conclusion}
In this paper, we propose FastGL, a GPU-efficient GNN training framework for large-scale 
graphs that 
simultaneously accelerates the memory IO, computation, and sample phases. 
Specifically, we
propose the 
Match-Reorder strategy to accelerate the memory IO by reusing 
the overlapping nodes between different mini-batches
without requiring any
extra GPU memory, which
is well suited to the 
training on large-scale graphs. 
Moreover, we put forward the Memory-Aware computation
method, 
which 
redesigns the memory access pattern during the computation
to increase the overall bandwidth
utilization, thus improving the 
GPU performance and 
accelerating the computation phase. 
Finally, we also present a 
Fused-Map sampling approach to 
avoid the thread synchronizations and accelerate the sample phase 
through a fused mechanism.
Extensive experiments compared with the prior art 
demonstrate the superiority 
of our FastGL. 
\endgroup

\section{Acknowledgments}
We extend our gratitude to the the ASPLOS reviewers 
who participated in the evaluation of FastGL for their valuable insights and feedback.
This work was supported in part by the STI 2030-Major Projects 
(No.2021ZD0201504), the Jiangsu Key Research and Development Plan (No.BE2021012-2),
the National Natural Science Foundation of China (No.62106267), the Key Research 
and Development Program of Jiangsu Province (Grants No.BE2023016) and 
the Beijing Municipal Science and Technology Project (No.Z231100010323002).

\bibliographystyle{plain}
\bibliography{references}

\begin{thebibliography}{10}

\bibitem{bojchevski2020scaling}
Aleksandar Bojchevski, Johannes Gasteiger, Bryan Perozzi, Amol Kapoor, Martin
  Blais, Benedek R{\'o}zemberczki, Michal Lukasik, and Stephan G{\"u}nnemann.
\newblock Scaling graph neural networks with approximate pagerank.
\newblock In {\em Proceedings of the 26th ACM SIGKDD International Conference
  on Knowledge Discovery \& Data Mining}, pages 2464--2473, 2020.
\newblock
  doi:{\href{https://doi.org/10.1145/3394486.3403296}{\textcolor{red}{https://doi.org/10.1145/3394486.3403296}}}.

\bibitem{cai2023dsp}
Zhenkun Cai, Qihui Zhou, Xiao Yan, Da~Zheng, Xiang Song, Chenguang Zheng, James
  Cheng, and George Karypis.
\newblock Dsp: Efficient gnn training with multiple gpus.
\newblock In {\em Proceedings of the 28th ACM SIGPLAN Annual Symposium on
  Principles and Practice of Parallel Programming}, pages 392--404, 2023.
\newblock
  doi:{\href{https://doi.org/10.1145/3572848.3577528}{\textcolor{red}{https://doi.org/10.1145/3572848.3577528}}}.

\bibitem{chen2018stochastic}
Jianfei Chen, Jun Zhu, and Le~Song.
\newblock Stochastic training of graph convolutional networks with variance
  reduction.
\newblock In {\em International Conference on Machine Learning}, pages
  942--950. PMLR, 2018.
\newblock
  doi:{\href{https://doi.org/10.48550/arXiv.1710.10568}{\textcolor{red}{https://doi.org/10.48550/arXiv.1710.10568}}}.

\bibitem{chen2018fastgcn}
Jie Chen, Tengfei Ma, and Cao Xiao.
\newblock Fastgcn: Fast learning with graph convolutional networks via
  importance sampling.
\newblock In {\em International Conference on Learning Representations}, 2018.
\newblock
  doi:{\href{https://doi.org/10.48550/arXiv.1801.10247}{\textcolor{red}{https://doi.org/10.48550/arXiv.1801.10247}}}.

\bibitem{chiang2019cluster}
Wei-Lin Chiang, Xuanqing Liu, Si~Si, Yang Li, Samy Bengio, and Cho-Jui Hsieh.
\newblock Cluster-gcn: An efficient algorithm for training deep and large graph
  convolutional networks.
\newblock In {\em Proceedings of the 25th ACM SIGKDD international conference
  on knowledge discovery \& data mining}, pages 257--266, 2019.
\newblock
  doi:{\href{https://doi.org/10.1145/3292500.3330925}{\textcolor{red}{https://doi.org/10.1145/3292500.3330925}}}.

\bibitem{ProgrammingGuide}
NVIDIA Corporation.
\newblock Cuda c++ programming guide,
  https://docs.nvidia.com/cuda/cuda-c-programming-guide/index.html, 2023.

\bibitem{nccl}
NVIDIA Corporation.
\newblock Nvidia collective communications library (nccl), 2023.

\bibitem{hopper}
NVIDIA Corporation.
\newblock Nvidia grace hopper superchip architecture whitepaper,
  https://resources.nvidia.com/en-us-grace-cpu/nvidia-grace-hopper, 2023.

\bibitem{fan2019graph}
Wenqi Fan, Yao Ma, Qing Li, Yuan He, Eric Zhao, Jiliang Tang, and Dawei Yin.
\newblock Graph neural networks for social recommendation.
\newblock In {\em The world wide web conference}, pages 417--426, 2019.
\newblock
  doi:{\href{https://doi.org/10.1145/3308558.3313488}{\textcolor{red}{https://doi.org/10.1145/3308558.3313488}}}.

\bibitem{fey2019fast}
Matthias Fey and Jan~Eric Lenssen.
\newblock Fast graph representation learning with pytorch geometric.
\newblock {\em arXiv preprint arXiv:1903.02428}, 2019.
\newblock
  doi:{\href{https://doi.org/10.48550/arXiv.1903.02428}{\textcolor{red}{https://doi.org/10.48550/arXiv.1903.02428}}}.

\bibitem{gandhi2021p3}
Swapnil Gandhi and Anand~Padmanabha Iyer.
\newblock P3: Distributed deep graph learning at scale.
\newblock In {\em 15th $\{$USENIX$\}$ Symposium on Operating Systems Design and
  Implementation ($\{$OSDI$\}$ 21)}, pages 551--568, 2021.

\bibitem{hamilton2017inductive}
Will Hamilton, Zhitao Ying, and Jure Leskovec.
\newblock Inductive representation learning on large graphs.
\newblock {\em Advances in neural information processing systems}, 30, 2017.
\newblock
  doi:{\href{https://dl.acm.org/doi/10.5555/3294771.3294869}{\textcolor{red}{https://dl.acm.org/doi/10.5555/3294771.3294869}}}.

\bibitem{hu2020open}
Weihua Hu, Matthias Fey, Marinka Zitnik, Yuxiao Dong, Hongyu Ren, Bowen Liu,
  Michele Catasta, and Jure Leskovec.
\newblock Open graph benchmark: Datasets for machine learning on graphs.
\newblock {\em Advances in neural information processing systems},
  33:22118--22133, 2020.
\newblock
  doi:{\href{https://dl.acm.org/doi/10.5555/3495724.3497579}{\textcolor{red}{https://dl.acm.org/doi/10.5555/3495724.3497579}}}.

\bibitem{huan2022t}
Chengying Huan, Shuaiwen~Leon Song, Yongchao Liu, Heng Zhang, Hang Liu, Charles
  He, Kang Chen, Jinlei Jiang, and Yongwei Wu.
\newblock T-gcn: A sampling based streaming graph neural network system with
  hybrid architecture.
\newblock In {\em Proceedings of the International Conference on Parallel
  Architectures and Compilation Techniques}, pages 69--82, 2022.
\newblock
  doi:{\href{https://doi.org/10.1145/3559009.3569648}{\textcolor{red}{https://doi.org/10.1145/3559009.3569648}}}.

\bibitem{huang2018adaptive}
Wenbing Huang, Tong Zhang, Yu~Rong, and Junzhou Huang.
\newblock Adaptive sampling towards fast graph representation learning.
\newblock {\em Advances in neural information processing systems}, 31, 2018.
\newblock
  doi:{\href{https://dl.acm.org/doi/abs/10.5555/3327345.3327367}{\textcolor{red}{https://dl.acm.org/doi/abs/10.5555/3327345.3327367}}}.

\bibitem{jangda2021accelerating}
Abhinav Jangda, Sandeep Polisetty, Arjun Guha, and Marco Serafini.
\newblock Accelerating graph sampling for graph machine learning using gpus.
\newblock In {\em Proceedings of the Sixteenth European Conference on Computer
  Systems}, pages 311--326, 2021.
\newblock
  doi:{\href{https://doi.org/10.1145/3447786.3456244}{\textcolor{red}{https://doi.org/10.1145/3447786.3456244}}}.

\bibitem{jia2020improving}
Zhihao Jia, Sina Lin, Mingyu Gao, Matei Zaharia, and Alex Aiken.
\newblock Improving the accuracy, scalability, and performance of graph neural
  networks with roc.
\newblock {\em Proceedings of Machine Learning and Systems}, 2:187--198, 2020.

\bibitem{jin2020multi}
Bowen Jin, Chen Gao, Xiangnan He, Depeng Jin, and Yong Li.
\newblock Multi-behavior recommendation with graph convolutional networks.
\newblock In {\em Proceedings of the 43rd International ACM SIGIR Conference on
  Research and Development in Information Retrieval}, pages 659--668, 2020.
\newblock
  doi:{\href{https://doi.org/10.1145/3397271.3401072}{\textcolor{red}{https://doi.org/10.1145/3397271.3401072}}}.

\bibitem{kaler2022accelerating}
Tim Kaler, Nickolas Stathas, Anne Ouyang, Alexandros-Stavros Iliopoulos, Tao
  Schardl, Charles~E Leiserson, and Jie Chen.
\newblock Accelerating training and inference of graph neural networks with
  fast sampling and pipelining.
\newblock {\em Proceedings of Machine Learning and Systems}, 4:172--189, 2022.

\bibitem{khatua2023igb}
Arpandeep Khatua, Vikram~Sharma Mailthody, Bhagyashree Taleka, Tengfei Ma,
  Xiang Song, and Wen-mei Hwu.
\newblock Igb: Addressing the gaps in labeling, features, heterogeneity, and
  size of public graph datasets for deep learning research.
\newblock In {\em Proceedings of the 29th ACM SIGKDD Conference on Knowledge
  Discovery and Data Mining}, pages 4284--4295, 2023.

\bibitem{kipf2016semi}
Thomas~N Kipf and Max Welling.
\newblock Semi-supervised classification with graph convolutional networks.
\newblock In {\em International Conference on Learning Representations}, 2016.
\newblock
  doi:{\href{https://doi.org/10.48550/arXiv.1609.02907}{\textcolor{red}{https://doi.org/10.48550/arXiv.1609.02907}}}.

\bibitem{klimke2022cooperative}
Marvin Klimke, Benjamin V{\"o}lz, and Michael Buchholz.
\newblock Cooperative behavior planning for automated driving using graph
  neural networks.
\newblock In {\em 2022 IEEE Intelligent Vehicles Symposium (IV)}, pages
  167--174. IEEE, 2022.
\newblock
  doi:{\href{https://doi.org/10.1109/IV51971.2022.9827230}{\textcolor{red}{https://doi.org/10.1109/IV51971.2022.9827230}}}.

\bibitem{lerer2019pytorch}
Adam Lerer, Ledell Wu, Jiajun Shen, Timothee Lacroix, Luca Wehrstedt, Abhijit
  Bose, and Alex Peysakhovich.
\newblock Pytorch-biggraph: A large scale graph embedding system.
\newblock {\em Proceedings of Machine Learning and Systems}, 1:120--131, 2019.
\newblock
  doi:{\href{https://doi.org/10.48550/arXiv.1903.12287}{\textcolor{red}{https://doi.org/10.48550/arXiv.1903.12287}}}.

\bibitem{li2021gcnax}
Jiajun Li, Ahmed Louri, Avinash Karanth, and Razvan Bunescu.
\newblock Gcnax: A flexible and energy-efficient accelerator for graph
  convolutional neural networks.
\newblock In {\em 2021 IEEE International Symposium on High-Performance
  Computer Architecture (HPCA)}, pages 775--788. IEEE, 2021.
\newblock
  doi:{\href{https://doi.org/10.1109/HPCA51647.2021.00070}{\textcolor{red}{https://doi.org/10.1109/HPCA51647.2021.00070}}}.

\bibitem{lin2020pagraph}
Zhiqi Lin, Cheng Li, Youshan Miao, Yunxin Liu, and Yinlong Xu.
\newblock Pagraph: Scaling gnn training on large graphs via computation-aware
  caching.
\newblock In {\em Proceedings of the 11th ACM Symposium on Cloud Computing},
  pages 401--415, 2020.
\newblock
  doi:{\href{https://doi.org/10.1145/3419111.3421281}{\textcolor{red}{https://doi.org/10.1145/3419111.3421281}}}.

\bibitem{liu2023bgl}
Tianfeng Liu, Yangrui Chen, Dan Li, Chuan Wu, Yibo Zhu, Jun He, Yanghua Peng,
  Hongzheng Chen, Hongzhi Chen, and Chuanxiong Guo.
\newblock $\{$BGL$\}$:$\{$GPU-Efficient$\}$$\{$GNN$\}$ training by optimizing
  graph data $\{$I/O$\}$ and preprocessing.
\newblock In {\em 20th USENIX Symposium on Networked Systems Design and
  Implementation (NSDI 23)}, pages 103--118, 2023.
\newblock
  doi:{\href{https://doi.org/10.48550/arXiv.2112.08541}{\textcolor{red}{https://doi.org/10.48550/arXiv.2112.08541}}}.

\bibitem{ma2019neugraph}
Lingxiao Ma, Zhi Yang, Youshan Miao, Jilong Xue, Ming Wu, Lidong Zhou, and
  Yafei Dai.
\newblock $\{$NeuGraph$\}$: Parallel deep neural network computation on large
  graphs.
\newblock In {\em 2019 USENIX Annual Technical Conference (USENIX ATC 19)},
  pages 443--458, 2019.

\bibitem{pandey2020c}
Santosh Pandey, Lingda Li, Adolfy Hoisie, Xiaoye~S Li, and Hang Liu.
\newblock C-saw: A framework for graph sampling and random walk on gpus.
\newblock In {\em SC20: International Conference for High Performance
  Computing, Networking, Storage and Analysis}, pages 1--15. IEEE, 2020.
\newblock
  doi:{\href{https://doi.org/10.1109/SC41405.2020.00060}{\textcolor{red}{https://doi.org/10.1109/SC41405.2020.00060}}}.

\bibitem{paszke2019pytorch}
Adam Paszke, Sam Gross, Francisco Massa, Adam Lerer, James Bradbury, Gregory
  Chanan, Trevor Killeen, Zeming Lin, Natalia Gimelshein, Luca Antiga, et~al.
\newblock Pytorch: An imperative style, high-performance deep learning library.
\newblock {\em Advances in neural information processing systems}, 32, 2019.
\newblock
  doi:{\href{https://dl.acm.org/doi/10.5555/3454287.3455008}{\textcolor{red}{https://dl.acm.org/doi/10.5555/3454287.3455008}}}.

\bibitem{perozzi2014deepwalk}
Bryan Perozzi, Rami Al-Rfou, and Steven Skiena.
\newblock Deepwalk: Online learning of social representations.
\newblock In {\em Proceedings of the 20th ACM SIGKDD international conference
  on Knowledge discovery and data mining}, pages 701--710, 2014.
\newblock
  doi:{\href{https://doi.org/10.1145/2623330.2623732}{\textcolor{red}{https://doi.org/10.1145/2623330.2623732}}}.

\bibitem{song2022rethinking}
Shihui Song and Peng Jiang.
\newblock Rethinking graph data placement for graph neural network training on
  multiple gpus.
\newblock In {\em Proceedings of the 36th ACM International Conference on
  Supercomputing}, pages 1--10, 2022.
\newblock
  doi:{\href{https://doi.org/10.1145/3503221.3508435}{\textcolor{red}{https://doi.org/10.1145/3503221.3508435}}}.

\bibitem{thorpe2021dorylus}
John Thorpe, Yifan Qiao, Jonathan Eyolfson, Shen Teng, Guanzhou Hu, Zhihao Jia,
  Jinliang Wei, Keval Vora, Ravi Netravali, Miryung Kim, et~al.
\newblock Dorylus: Affordable, scalable, and accurate $\{$GNN$\}$ training with
  distributed $\{$CPU$\}$ servers and serverless threads.
\newblock In {\em 15th USENIX Symposium on Operating Systems Design and
  Implementation (OSDI 21)}, pages 495--514, 2021.
\newblock
  doi:{\href{https://doi.org/10.48550/arXiv.2105.11118}{\textcolor{red}{https://doi.org/10.48550/arXiv.2105.11118}}}.

\bibitem{velivckovic2017graph}
Petar Veli{\v{c}}kovi{\'c}, Guillem Cucurull, Arantxa Casanova, Adriana Romero,
  Pietro Lio, and Yoshua Bengio.
\newblock Graph attention networks.
\newblock {\em arXiv preprint arXiv:1710.10903}, 2017.
\newblock
  doi:{\href{https://doi.org/10.48550/arXiv.1710.10903}{\textcolor{red}{https://doi.org/10.48550/arXiv.1710.10903}}}.

\bibitem{velivckovic2018graph}
Petar Veli{\v{c}}kovi{\'c}, Guillem Cucurull, Arantxa Casanova, Adriana Romero,
  Pietro Li{\`o}, and Yoshua Bengio.
\newblock Graph attention networks.
\newblock In {\em International Conference on Learning Representations}, 2018.
\newblock
  doi:{\href{https://doi.org/10.48550/arXiv.1710.10903}{\textcolor{red}{https://doi.org/10.48550/arXiv.1710.10903}}}.

\bibitem{wan2022bns}
Cheng Wan, Youjie Li, Ang Li, Nam~Sung Kim, and Yingyan Lin.
\newblock Bns-gcn: Efficient full-graph training of graph convolutional
  networks with partition-parallelism and random boundary node sampling.
\newblock {\em Proceedings of Machine Learning and Systems}, 4:673--693, 2022.
\newblock
  doi:{\href{https://doi.org/10.48550/arXiv.2203.10983}{\textcolor{red}{https://doi.org/10.48550/arXiv.2203.10983}}}.

\bibitem{wan2023scalable}
Xinchen Wan, Kaiqiang Xu, Xudong Liao, Yilun Jin, Kai Chen, and Xin Jin.
\newblock Scalable and efficient full-graph gnn training for large graphs.
\newblock {\em Proceedings of the ACM on Management of Data}, 1(2):1--23, 2023.
\newblock
  doi:{\href{https://doi.org/10.1145/3589288}{\textcolor{red}{https://doi.org/10.1145/3589288}}}.

\bibitem{wang2021flexgraph}
Lei Wang, Qiang Yin, Chao Tian, Jianbang Yang, Rong Chen, Wenyuan Yu, Zihang
  Yao, and Jingren Zhou.
\newblock Flexgraph: a flexible and efficient distributed framework for gnn
  training.
\newblock In {\em Proceedings of the Sixteenth European Conference on Computer
  Systems}, pages 67--82, 2021.
\newblock
  doi:{\href{https://doi.org/10.1145/3447786.3456229}{\textcolor{red}{https://doi.org/10.1145/3447786.3456229}}}.

\bibitem{wang2019deep}
Minjie~Yu Wang.
\newblock Deep graph library: Towards efficient and scalable deep learning on
  graphs.
\newblock In {\em ICLR workshop on representation learning on graphs and
  manifolds}, 2019.
\newblock
  doi:{\href{https://doi.org/10.48550/arXiv.1909.01315}{\textcolor{red}{https://doi.org/10.48550/arXiv.1909.01315}}}.

\bibitem{wang2021gnnadvisor}
Yuke Wang, Boyuan Feng, Gushu Li, Shuangchen Li, Lei Deng, Yuan Xie, and Yufei
  Ding.
\newblock $\{$GNNAdvisor$\}$: An adaptive and efficient runtime system for
  $\{$GNN$\}$ acceleration on $\{$GPUs$\}$.
\newblock In {\em 15th USENIX symposium on operating systems design and
  implementation (OSDI 21)}, pages 515--531, 2021.
\newblock
  doi:{\href{https://doi.org/10.48550/arXiv.2006.06608}{\textcolor{red}{https://doi.org/10.48550/arXiv.2006.06608}}}.

\bibitem{weng2020joint}
Xinshuo Weng, Ye~Yuan, and Kris Kitani.
\newblock Joint 3d tracking and forecasting with graph neural network and
  diversity sampling.
\newblock {\em arXiv preprint arXiv:2003.07847}, 2(6.2):1, 2020.
\newblock
  doi:{\href{https://doi.org/10.48550/arXiv.2003.07847}{\textcolor{red}{https://doi.org/10.48550/arXiv.2003.07847}}}.

\bibitem{xu2018powerful}
Keyulu Xu, Weihua Hu, Jure Leskovec, and Stefanie Jegelka.
\newblock How powerful are graph neural networks?
\newblock In {\em International Conference on Learning Representations}, 2018.
\newblock
  doi:{\href{https://doi.org/10.48550/arXiv.1810.00826}{\textcolor{red}{https://doi.org/10.48550/arXiv.1810.00826}}}.

\bibitem{zhu12aligraph}
Hongxia Yang.
\newblock Aligraph: A comprehensive graph neural network platform.
\newblock In {\em Proceedings of the 25th ACM SIGKDD international conference
  on knowledge discovery \& data mining}, pages 3165--3166, 2019.
\newblock
  doi:{\href{https://doi.org/10.1145/3292500.3340404}{\textcolor{red}{https://doi.org/10.1145/3292500.3340404}}}.

\bibitem{yang2022gnnlab}
Jianbang Yang, Dahai Tang, Xiaoniu Song, Lei Wang, Qiang Yin, Rong Chen,
  Wenyuan Yu, and Jingren Zhou.
\newblock Gnnlab: a factored system for sample-based gnn training over gpus.
\newblock In {\em Proceedings of the Seventeenth European Conference on
  Computer Systems}, pages 417--434, 2022.
\newblock
  doi:{\href{https://doi.org/10.1145/3492321.3519557}{\textcolor{red}{https://doi.org/10.1145/3492321.3519557}}}.

\bibitem{yang2023betty}
Shuangyan Yang, Minjia Zhang, Wenqian Dong, and Dong Li.
\newblock Betty: Enabling large-scale gnn training with batch-level graph
  partitioning.
\newblock In {\em Proceedings of the 28th ACM International Conference on
  Architectural Support for Programming Languages and Operating Systems, Volume
  2}, pages 103--117, 2023.
\newblock
  doi:{\href{https://doi.org/10.1145/3575693.3575725}{\textcolor{red}{https://doi.org/10.1145/3575693.3575725}}}.

\bibitem{yang2020hagerec}
Zuoxi Yang and Shoubin Dong.
\newblock Hagerec: Hierarchical attention graph convolutional network
  incorporating knowledge graph for explainable recommendation.
\newblock {\em Knowledge-Based Systems}, 204:106194, 2020.
\newblock
  doi:{\href{https://doi.org/10.1016/j.knosys.2020.106194}{\textcolor{red}{https://doi.org/10.1016/j.knosys.2020.106194}}}.

\bibitem{ying2018graph}
Rex Ying, Ruining He, Kaifeng Chen, Pong Eksombatchai, William~L Hamilton, and
  Jure Leskovec.
\newblock Graph convolutional neural networks for web-scale recommender
  systems.
\newblock In {\em Proceedings of the 24th ACM SIGKDD international conference
  on knowledge discovery \& data mining}, pages 974--983, 2018.
\newblock
  doi:{\href{https://doi.org/10.1145/3219819.3219890}{\textcolor{red}{https://doi.org/10.1145/3219819.3219890}}}.

\bibitem{zeng2019graphsaint}
Hanqing Zeng, Hongkuan Zhou, Ajitesh Srivastava, Rajgopal Kannan, and Viktor
  Prasanna.
\newblock Graphsaint: Graph sampling based inductive learning method.
\newblock In {\em International Conference on Learning Representations}, 2019.
\newblock
  doi:{\href{https://doi.org/10.48550/arXiv.1907.04931}{\textcolor{red}{https://doi.org/10.48550/arXiv.1907.04931}}}.

\bibitem{zhang2021biased}
Qingru Zhang, David Wipf, Quan Gan, and Le~Song.
\newblock A biased graph neural network sampler with near-optimal regret.
\newblock {\em Advances in Neural Information Processing Systems},
  34:8833--8844, 2021.
\newblock
  doi:{\href{https://doi.org/10.48550/arXiv.2103.01089}{\textcolor{red}{https://doi.org/10.48550/arXiv.2103.01089}}}.

\bibitem{zheng2022bytegnn}
Chenguang Zheng, Hongzhi Chen, Yuxuan Cheng, Zhezheng Song, Yifan Wu, Changji
  Li, James Cheng, Hao Yang, and Shuai Zhang.
\newblock Bytegnn: efficient graph neural network training at large scale.
\newblock {\em Proceedings of the VLDB Endowment}, 15(6):1228--1242, 2022.
\newblock
  doi:{\href{https://doi.org/10.14778/3514061.3514069}{\textcolor{red}{https://doi.org/10.14778/3514061.3514069}}}.

\bibitem{zheng2020distdgl}
Da~Zheng, Chao Ma, Minjie Wang, Jinjing Zhou, Qidong Su, Xiang Song, Quan Gan,
  Zheng Zhang, and George Karypis.
\newblock Distdgl: distributed graph neural network training for billion-scale
  graphs.
\newblock In {\em 2020 IEEE/ACM 10th Workshop on Irregular Applications:
  Architectures and Algorithms (IA3)}, pages 36--44. IEEE, 2020.
\newblock
  doi:{\href{https://doi.org/10.1109/IA351965.2020.00011}{\textcolor{red}{https://doi.org/10.1109/IA351965.2020.00011}}}.

\bibitem{zou2019layer}
Difan Zou, Ziniu Hu, Yewen Wang, Song Jiang, Yizhou Sun, and Quanquan Gu.
\newblock Layer-dependent importance sampling for training deep and large graph
  convolutional networks.
\newblock {\em Advances in neural information processing systems}, 32, 2019.
\newblock
  doi:{\href{https://dl.acm.org/doi/10.5555/3454287.3455296}{\textcolor{red}{https://dl.acm.org/doi/10.5555/3454287.3455296}}}.

\end{thebibliography}

\end{document}